\documentclass[9pt,twocolumn,twoside]{osajnl}

\usepackage{times}

\usepackage{amsmath}
\usepackage{amssymb}
\usepackage{bm}
\usepackage{multirow}
\usepackage{mathtools}
\usepackage{units}
\usepackage{subfloat}
\usepackage{float}
\usepackage[]{units}
\usepackage{xcolor}
\usepackage{url}
\usepackage{bbm}
\usepackage{algorithm}
\usepackage{algpseudocode}
\usepackage{pbox}
\usepackage{placeins}
\usepackage{siunitx}
\usepackage{tabularx} %
\usepackage{rotating}
\usepackage{subcaption}
\usepackage{appendix}

\usepackage[TS1,T1]{fontenc}
\DeclareSIUnit[number-unit-product = {}]{\inch}{\textquotedbl}

\usepackage{tikz,pgfplots}
\usetikzlibrary{matrix}
\pgfplotsset{compat=1.15}
\usepackage{etex} 
\usetikzlibrary{arrows,automata}
\usetikzlibrary{positioning}
\usepgfplotslibrary{groupplots}
\usepackage{tikz-3dplot}
\tikzset{
	state/.style={
		rectangle,
		rounded corners,
		draw=black, very thick,
		minimum height=2em,
		inner sep=2pt,
		text centered,
	},
}
\tikzset{
	info/.style={
		rectangle,
		draw=black, thin,
		minimum height=2em,
		inner sep=2pt,
		text centered,
	},
}
\usetikzlibrary{positioning}
\usetikzlibrary{calc}
\usetikzlibrary{arrows,shapes,backgrounds}

\definecolor{ours}{rgb}{0.4660, 0.6740, 0.1880}
\definecolor{fastplanner}{rgb}{0.85,0.325,0.098}
\definecolor{reactive}{rgb}{0,0.447,0.741}
\definecolor{blind}{rgb}{1.0, 0.0, 0.0}
\definecolor{m_yellow}{rgb}{0.929,0.694,0.125}
\definecolor{m_purple}{rgb}{0.4940, 0.1840, 0.5560}

\newcommand{\real}[0]{{\rm I\!R}}
\newcommand{\nat}[0]{\mathbb{N}}

\newcommand{\traj}{\bm{\tau}}

\newcommand{\timevec}{\bm{\mathrm{T}}}

\newcommand{\ie}{\emph{i.e.~}}
\newcommand{\eg}{\emph{e.g.~}}

\DeclareMathOperator*{\minimize}{minimize}

\def\trans{^{\top}}

\journal{ol} 

\setboolean{shortarticle}{false}

\title{Learning High-Speed Flight in the Wild}

\author[1,$\dagger$,*]{Antonio Loquercio}
\author[1,$\dagger$]{Elia Kaufmann}
\author[2]{Ren\'{e} Ranftl}
\author[2]{Matthias M\"uller}
\author[3]{\\Vladlen Koltun}
\author[1]{Davide Scaramuzza}

\affil[1]{Robotics and Perception Group, UZH, Zurich, Switzerland}
\affil[2]{Intelligent Systems Lab, Intel, Munich, Germany}
\affil[3]{Intelligent Systems Lab, Intel, Santa Clara, CA, USA}
\affil[$\dagger$]{These authors contributed equally to this work}
\affil[*]{Corresponding author: loquercio@ifi.uzh.ch}

\dates{This is the accepted version of Science Robotics Vol. 6, Issue 59, abg5810 (2021) \\ DOI: 10.1126/scirobotics.abg5810}

\begin{abstract}
Quadrotors are agile. Unlike most other machines, they can traverse extremely complex environments at high speeds. To date, only expert human pilots have been able to fully exploit their capabilities. Autonomous operation with onboard sensing and computation has been limited to low speeds.
State-of-the-art methods generally separate the navigation problem into subtasks: sensing, mapping, and planning.
Although this approach has proven successful at low speeds, the separation it builds upon can be problematic for high-speed navigation in cluttered environments.
The subtasks are executed sequentially, leading to increased processing latency and a compounding of errors through the pipeline.
Here we propose an end-to-end approach that can autonomously fly quadrotors through complex natural and human-made environments at high speeds, with purely onboard sensing and computation.
The key principle is to directly map noisy sensory observations to collision-free trajectories in a receding-horizon fashion.
This direct mapping drastically reduces processing latency and increases robustness to noisy and incomplete perception.
The sensorimotor mapping is performed by a convolutional network that is trained \emph{exclusively} in simulation via privileged learning: imitating an expert with access to privileged information.
By simulating realistic sensor noise, our approach achieves zero-shot transfer from simulation to challenging real-world environments that were never experienced during training: dense forests, snow-covered terrain, derailed trains, and collapsed buildings.
Our work demonstrates that end-to-end policies trained in simulation enable high-speed autonomous flight through challenging environments, outperforming traditional obstacle avoidance pipelines.
\end{abstract}

\begin{document}

\maketitle
\begin{figure*}
    \centering
    \includegraphics[width=\textwidth]{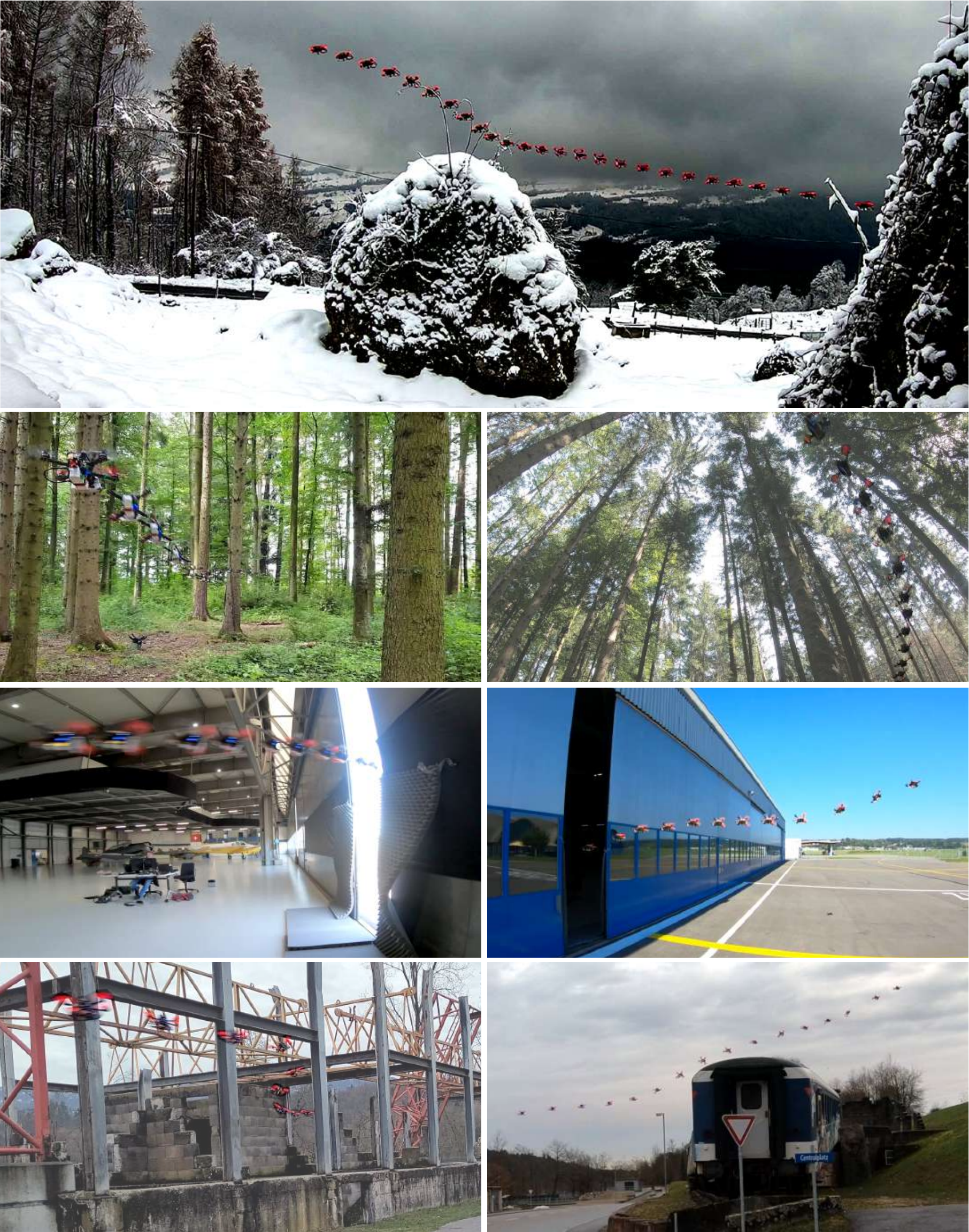}
    \caption{\textbf{Time-lapse illustrations of agile navigation} at speeds between 5 and 10 m/s in a variety of environments. From top to bottom: A mountain trail, a forest, an airplane hangar, and a disaster zone. The middle rows show the same environments, but from different viewpoints. The last row shows navigation through a collapsed building and above a derailed train. Please watch \href{https://www.youtube.com/watch?v=m89bNn6RFoQ}{Movie 1} to get a better sense of the speed and agility of our approach.
    \label{fig:1}}
\end{figure*}

\section*{Code and Multimedia Material}
A video demonstrating our approach is available at \url{https://youtu.be/m89bNn6RFoQ}. We publicly release code and training datasets at \url{https://github.com/uzh-rpg/agile_autonomy}.

\section{Introduction} \label{sec:introduction}

Quadrotors are among the most agile and dynamic machines ever created\footnote{The quadrotor used for the experiments in this paper has a maximum acceleration of 4g. Formula 1 cars achieve accelerations of up to 1.45g, and the Eurofighter Typhoon reaches a longitudinal acceleration of up to 1.15g.}~\cite{Verbeke18IJMAV,AcrobaticDrones}.
Thanks to their agility, they can traverse complex environments, ranging from cluttered forests to urban canyons, and reach locations that are otherwise inaccessible to humans and machines alike.
This ability has led to their application in fields such as search and rescue, logistics, security, infrastructure, entertainment, and agriculture~\cite{drone_commercial}.
In the majority of these existing applications, the quadrotor needs to be controlled by expert human pilots, who take years to train, and are thus an expensive and scarce resource.
Infusing quadrotors with autonomy, that is the capability to safely operate in the world without the need for human intervention, has the potential to massively enhance their usefulness and 
to revolutionize whole industries.
However, the development of autonomous quadrotors that can navigate in complex environments with the agility and safety of expert human pilots or birds is a long-standing challenge that is still open.

The limiting factor for autonomous agile flight in arbitrary unknown environments is the coupling of fast and robust perception with effective planning.
The perception system has to be robust to disturbances such as sensor noise,
motion blur, and changing illumination conditions.
In addition, an effective planner is necessary to find a path that is both dynamically feasible and collision-free while relying only on noisy and partial observations of the environment.
These requirements, together with the limited computational resources that are available onboard, make it difficult to achieve reliable perception and planning at low latency and high speeds~\cite{Falanga19ral_howfast}.

Various approaches to enable autonomous flight have been proposed in the literature. Some works tackle only perception and build high-quality maps from imperfect measurements~\cite{Heng14jfr,Saeedi17Us,Scaramuzza14RAM,Faessler16JFR,BloschWSS10}, whereas others focus on planning without considering perception errors~\cite{Bry2011icra,Richter2013ISRR,allen2016real,liu2018search}.
Numerous systems that combine online mapping with traditional planning algorithms have been proposed to achieve autonomous flight in previously unknown environments~\cite{Oleynikova17IROS,Ozaslan17RAL,Mohta18jfr,Barry18jfr,zhou2019robust,ryll2019efficient,Tordesillas19Iros,Cieslewsk17iros,Zhang18icra}.
A taxonomy of prior works is presented in Figure~\ref{fig:pc:map_rel_work} in the Supplementary Materials.

The division of the navigation task into the mapping and planning subtasks is attractive from an engineering perspective, because it enables parallel progress on each component and makes the overall system interpretable.
However, it leads to pipelines that largely neglect interactions between the different stages and thus compound errors~\cite{Zhang18icra}.
Their sequential nature also introduces additional latency, making high-speed and agile maneuvers difficult to impossible~\cite{Falanga19ral_howfast}.
While these issues can be mitigated to some degree by careful hand-tuning and engineering, the divide-and-conquer principle that 
has been prevalent in research on autonomous flight in unknown environments for many years imposes fundamental limits on the speed and agility that a robotic system can achieve~\cite{Loianno20JFR}.

In contrast to these traditional pipelines, some recent works propose to learn end-to-end policies directly from data without explicit mapping and planning stages~\cite{Ross13icra,SadeghiL17Rss,Gandhi17Iros,Loquercio18ral}.
These policies are trained by imitating a human~\cite{Ross13icra,Loquercio18ral}, from experience that was collected in simulation~\cite{SadeghiL17Rss}, or directly in the real world~\cite{Gandhi17Iros}.
Because the number of samples required to train general navigation policies is very high,
existing approaches impose constraints on the quadrotor's motion model, for example by constraining the platform to planar motion~\cite{Loquercio18ral,Gandhi17Iros,Ross13icra} and/or discrete actions~\cite{SadeghiL17Rss}, at the cost of reduced maneuverability and agility.
More recent work has demonstrated that very agile control policies can be trained in simulation~\cite{kaufmann2020RSS}.
Policies produced by the last approach can successfully perform acrobatic maneuvers, 
but can only operate in unobstructed free space and are essentially blind to obstacles in the environment.

Here we present an approach to fly a quadrotor at high speeds in a variety of environments with complex obstacle geometry (Figure~\ref{fig:1} and \href{https://www.youtube.com/watch?v=m89bNn6RFoQ}{Movie 1}) while having access to only onboard sensing and computation.
By predicting navigation commands directly from sensor measurements,
we decrease the latency between perception and action while simultaneously being robust to perception artifacts, such as motion blur, missing data, and sensor noise.
To deal with sample complexity and not endanger the physical platform, we train the policy \emph{exclusively} in simulation.
We leverage abstraction of the input data to transfer the policy from simulation to reality~\cite{mueller2018abstraction,kaufmann2020RSS}.
To this end, we utilize a stereo matching algorithm to provide depth images as input to the policy.
We show that this representation is both rich enough to safely navigate through complex environments and abstract enough to bridge simulation and reality. 
Our choice of input representation guarantees a strong similarity of the noise models between simulated and real observations and gives our policy robustness against common perceptual artifacts in existing depth sensors.

We train the navigation policy via privileged learning~\cite{chen2020learning} on demonstrations that are provided by a sampling-based expert.
Our expert has privileged access to a representation of the environment in the form of a three-dimensional (3D) point cloud as well as perfect knowledge about the state of the quadrotor. %
Since simulation does not impose real-time constraints, the expert additionally has an unconstrained computational budget.
While existing global planning algorithms~\cite{Bry2011icra,allen2016real,liu2018search} generally output a single trajectory, 
our expert uses Metropolis-Hastings (M-H) sampling to compute a \textit{distribution} of collision-free trajectories. This 
captures the multi-modal nature of the navigation task where many equally valid solutions can exist (for example, going either left or right around an obstacle).
We therefore use our planner to compute trajectories with a short time horizon to ensure that they are predictable from onboard 
sensors and that the sampler remains computationally tractable. We bias the sampler toward obstacle-free regions by 
conditioning it on trajectories from a classic global planning algorithm~\cite{liu2018search}.

We also reflect the multi-modal nature of the problem in the design and training of the neural network policy. Our policy takes a noisy depth image and inertial measurements as sensory inputs and produces a set of short-term trajectories together
with an estimate of individual trajectory costs. The trajectories are represented as high-order polynomials to ensure
dynamical feasibility. We train the policy using a multi-hypothesis winner-takes-all loss that adaptively maps the predicted trajectories to the best trajectories that have been found by the sampling-based expert. At test time, we use the predicted
trajectory costs to decide which trajectory to execute in a receding horizon. The policy network is designed to be extremely lightweight,
which ensures that it can be executed onboard the quadrotor at the update rates required for high-speed flight.

\begin{figure*}
    \centering
  \includegraphics[width=\linewidth]{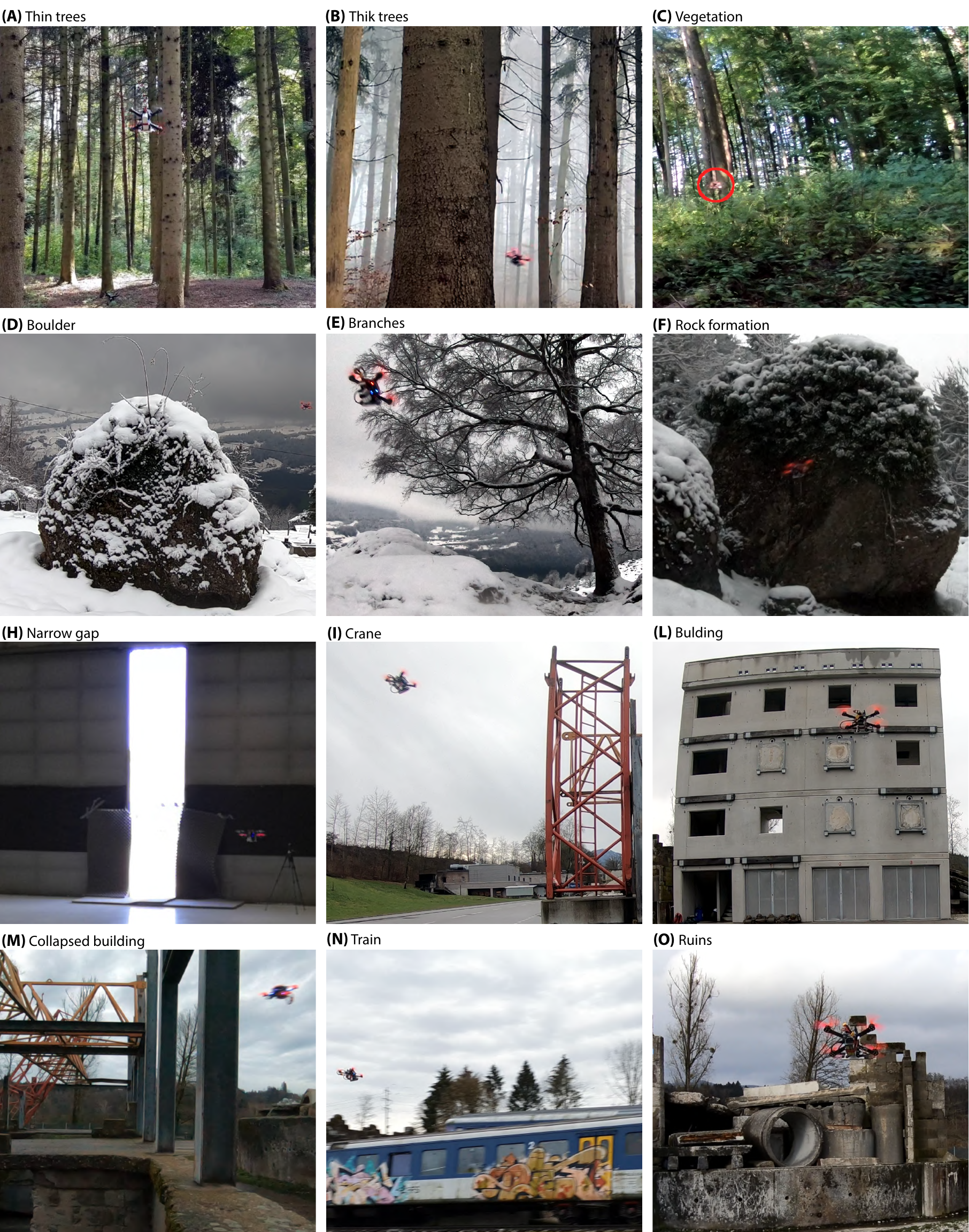}
    \caption{\textbf{Testing Environments.} Zero-shot generalization of our approach in complex natural (\textbf{A} to \textbf{F}) and human-made (\textbf{H} to \textbf{O}) environments. The encountered obstacles can often be avoided in multiple directions and have very different size and structure.
}\label{fig:environments}
\end{figure*}

The resulting policy can fly a physical quadrotor in natural and human-made environments at speeds that are unreachable by existing methods.
We achieve this in a zero-shot generalization setting: we train on randomly generated obstacle courses composed of simple off-the-shelf objects, such as schematic trees and a small set of convex shapes such as cylinders and cubes.
We then directly deploy the policy in the physical world without any adaptation or fine-tuning.
Our platform experiences conditions at test time that were never seen during training. Examples include high dynamic range (when flying from indoor environments to outdoor environments), poorly textured surfaces (indoor environments and snow-covered terrain), thick vegetation in forests, and the irregular and complex layout of a disaster scenario (Figure~\ref{fig:environments}).
These results suggest that our methodology enables a multitude of applications that rely on agile autonomous drones with purely onboard sensing and computation.

\section{Results} \label{sec:experiments}

Our experiments in simulation show that the proposed approach reduces the failure rate up to 10 times with respect to state-of-the-art methods.
We confirm our results in a variety of real-world environments using a custom-built physical quadrotor; we deploy our policy trained in simulation without any further adaptations.
In all experiments, the drone was provided with a reference trajectory, which is not collision-free (Figure~\ref{fig:results_real_world}-C, depicted in red), to encode the intended flight path.
This reference can be provided by a user or a higher-level planning algorithm.
The drone is tasked to follow that flight path and make adjustments as necessary to avoid obstacles.
Recordings of the experiments can be found in \href{https://www.youtube.com/watch?v=m89bNn6RFoQ}{Movie~1}.

\subsection{High-Speed Flight in the Wild}

We tested our approach in diverse real-world environments, as illustrated in Figures~\ref{fig:1}~and~\ref{fig:environments}. %
High-speed flight in these environments is very challenging because of their complex structure (\eg thick vegetation, thin branches, or collapsed walls) and multiple options available to avoid obstacles.
In addition, a high-level understanding of the environment is necessary, for example to pass through far-away openings (Figure~\ref{fig:results_real_world}-C) or nearby obstacles (Figure~\ref{fig:environments}-M).
Flying at high speeds also necessitates low-latency perception and robustness to sensor noise, which is worsened by challenging illumination conditions and low-texture surfaces (\eg because of snow). 
At the same time, only limited computational resources are available onboard.
Despite all of these challenges, our approach was able to successfully navigate in all environments
that it was tested in.
Note that our policy was trained in simulation and was never exposed to any of these environments or conditions at training time.

We measure performance according to success rate, \emph{i.e.} the percentage of successful runs over the total number of runs, where we
consider a run successful if the drone reaches the goal location within a radius of \SI{5}{\meter} without crashing.
We performed a total of 56 experiments at different speeds.
Overall, our approach was always able to achieve the goal when not colliding into an obstacle.
We report the cumulative and individual success rates at various speeds in Figure~\ref{fig:results_real_world} (D and E).
Our experimental platform performs state estimation and depth perception onboard using vision-based sensors. 
A detailed description of the platform is available in section~\ref{sec:experimental_platform}.
We group the environments that were used for experiments into two classes, \textit{natural} and \textit{human-made}, and highlight the distinct challenges that these types of environments present for agile autonomous flight.

\subsection*{Natural environments}
\begin{figure*}
    \centering
     \makebox[\textwidth][c]{
    \includegraphics[width=\textwidth]{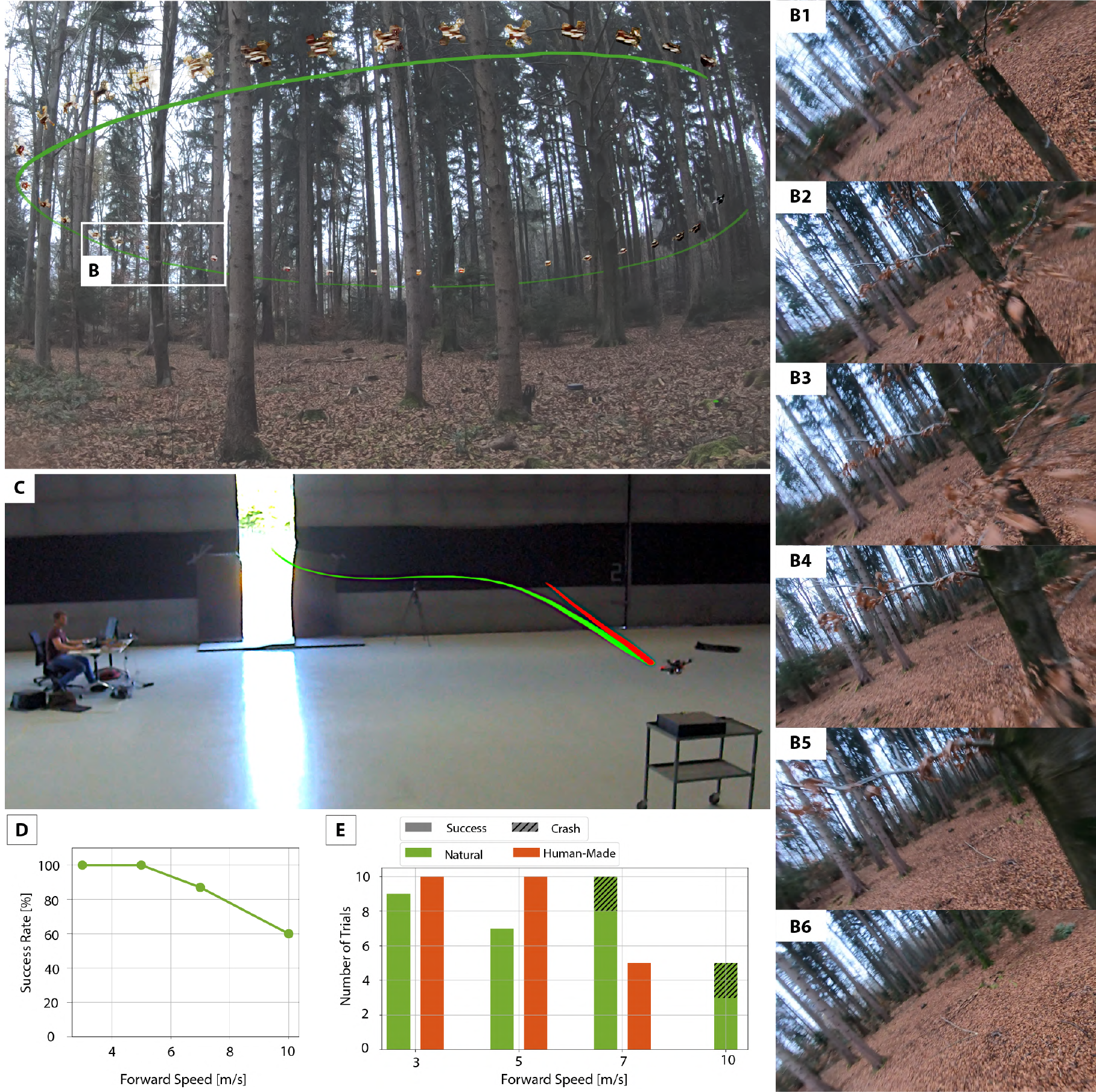}
    }
    \caption{\textbf{Evaluation in indoor and outdoor environments}. (\textbf{A}) A circular path in the forest at an average speed of \SI{7}{\meter\per\second}. (\textbf{B}) Sequence of first-person views from the maneuver in \textbf{A}, observed during the avoidance of tree branches. The maneuver requires fine and fast adaptations in height (\textbf{B2},\textbf{B5}) and attitude (\textbf{B3},\textbf{B4}) to avoid the vegetation. After the obstacle is passed, the drone accelerates in the direction of travel (\textbf{B6}). (\textbf{C}) Comparison of reference trajectory passing through a wall  (in red), to actual flight path (in green) in a airplane hangar. (\textbf{D}) Success rates for all experiments aggregated according to flight speed. (\textbf{E}) Number of trials per environment class at different speeds.
\label{fig:results_real_world}}
\end{figure*}

We performed experiments in diverse natural environments: forests of different types and densities and steep snowy mountain terrains. 
Figures~\ref{fig:environments} (A and F) illustrate the heterogeneity of those environments. %
We performed experiments with two different reference trajectories: a \SI{40}{\meter}-long straight line and a circle with a \SI{6}{\meter} radius (Figure~\ref{fig:results_real_world}-A). 
Both trajectory types are not collision-free and would lead to a crash into obstacles if blindly executed.
We flew the straight line at different average speeds in the range of 3 to 10 $\SI{}{\meter\per\second}$.
Flying at these speeds requires very precise and low-latency control of position and attitude to avoid bushes and pass through the openings between trees and branches (Figure~\ref{fig:results_real_world}-B).
Traditional mapping and planning approaches generally fail to achieve high speeds in such conditions, as both the thick vegetation and the texture-less snow terrain often cause very noisy depth observations. %

We conducted a total of 31 experiments in natural environments.
At average speeds of 3 and 5 \SI{}{\meter\per\second} our approach consistently completed the task without experiencing a single crash.
For comparison, state-of-the-art methods with comparable actuation, sensing, and computation~\cite{Zhou19arxiv,zhou2019robust} achieve in similar environments a maximum average speed of \SI{2.29}{\meter\per\second}.
To study the limit of the system, we set the platform's average speed to \SI{7}{\meter\per\second}.
Despite the very high-speed, the maneuver was successfully completed in 8 out of 10 experiments.
The two failures happened when objects entered the field of view very late because of the high angular velocity of the platform.
Given the good performance at \SI{7}{\meter\per\second}, we push the average flight speed even higher to \SI{10}{\meter\per\second}.
At this speed, external disturbances, \eg aerodynamics, battery power drops, and motion-blur start to play an important role and widen the simulation to reality gap.
Nonetheless, we achieve a success rate of $60\%$, with failures mainly happening in the proximity of narrow openings less than a meter wide, where a single wrong action results in a crash.

\subsection*{Human-made environments}
We also tested our approach in a set of human-made environments, illustrated in Figure~\ref{fig:environments} (G to O). 
In these environments, the drone faces a different set of challenges. It has to avoid obstacles with a variety of sizes and shapes (\eg a train, a crane, a building, and ruins),  slalom through concrete structures (c.f. Figure~\ref{fig:environments}M), and exit a building through a single narrow opening (c.f. Figure~\ref{fig:environments}H).
The irregular and/or large structure of the encountered obstacles, the limited number of flyable openings, and the requirement to initiate the avoidance maneuver well in advance, offer a complementary set of challenges with respect to our natural testing environments. 

As in the natural environments, we provide the drone with a straight reference trajectory with length of \SI{40}{\meter}.
The reference trajectory is in direct collision with obstacles and its blind execution would result in a crash.
We performed a total of 19 experiments with flight speeds in the range of 3 to 7 $\SI{}{\meter\per\second}$.
Given the lower success rate experienced in the forest environment at \SI{10}{\meter\per\second}, we did not test at higher speeds to avoid fatal crashes.
As shown in Figure~\ref{fig:results_real_world}E, our approach is robust to zero-shot deployment in these environments, whose characteristics were never observed at training time, and consistently completes the task without crashing.

We further compare our approach to a commercial drone\footnote{The commercial drone (a Skydio~R1) was tasked to follow a person running through the narrow gap. The speed of this drone  cannot be enforced nor controlled, and was therefore estimated in post-processing.}.
Specifically, the drones are required to exit a hangar by passing through a narrow gap of about \SI{0.8}{\meter} in width (Figure~\ref{fig:results_real_world}-C).
At the start of the experiment, the drones are placed at about \SI{10}{\meter} in front of and about \SI{5}{\meter} to the right of the gap.
The task was represented by a straight reference trajectory passing through the wall (Figure~\ref{fig:results_real_world}C, in red).
This experiment is challenging since it requires a high-level understanding of the environment to turn early enough toward the gap.
The commercial drone, flying at a speed of about \SI{2.7}{\meter\per\second} consistently failed to pass through the gap across three experiments.
In two of the experiments it deemed the gap to be too small and stopped in front of it; in the third, it crashed into the wall.
In contrast, the low latency and robustness to perception failures of our approach enabled the drone to successfully fly through the narrow gap every time.
We performed a total of six experiments at flight speeds of 3 and 5 \SI{}{\meter\per\second} and never experienced a crash.

\subsection{Controlled experiments} \label{sec:sota_comp}
We perform a set of controlled experiments in simulation to compare the performance of our approach with several baselines.
We select two representative state-of-the-art approaches as baselines for navigation in unknown environments:
the mapping and planning method of Zhou et al.~\cite{zhou2019robust} (\emph{FastPlanner}) and the reactive planner of Florence et al.~\cite{florence2020integrated} (\emph{Reactive}).
The first approach~\cite{zhou2019robust} incrementally builds a map from a series of observations and plans a trajectory in this map to reach the goal while avoiding obstacles.
In contrast, the second approach~\cite{florence2020integrated} does not build a map but uses instantaneous depth information to select the best trajectory from a set of pre-defined motion primitives based on a cost that encodes collision and progress toward the goal.
The baselines receive the same input as our policy: the state of the platform, depth measurements from the stereo camera, and a goal in the form of a reference state that lies \SI{5}{\second} in the future.
To provide a notion of the difficulty of the environments, we additionally show a naive baseline that blindly follows the reference trajectory to the goal without avoiding obstacles (\emph{Blind}).
As in the real-world experiments, we compare the different approaches according to their success rate, which measures how often the drone reaches the goal location within a radius of 5 \SI{}{\meter} without crashing.
We perform all experiments in the Flightmare simulator~\cite{yunlong2020flightmare} using the RotorS~\cite{furrer2016rotors} Gazebo plugin for accurate physics modeling and Unity as a rendering engine~\cite{juliani2018unity}.
The experiments are conducted in four different environments: a forest, a narrow gap, a disaster scenario, and a city street.
Those environments offer different challenges: from high-density but simple obstacle geometry in the forest, to complex obstacle geometries in the urban scenarios.
Sample observations collected from these environments are available in Figure~\ref{fig:sim_environments} and \href{https://youtu.be/ALs60ij8JA8}{Movie S2}.
These environments are available out-of-the-box from the \href{https://assetstore.unity.com/}{Unity Asset store}.

The results of these experiments are summarized in Figure~\ref{fig:sim_results}.
Throughout all environments, a similar pattern can be observed.
At low speeds (\SI{3}{\meter\per\second}) all methods perform similarly.
However, as the speed increases, the baselines' performances quickly drop: already at \SI{5}{\meter\per\second}, no baseline is able to complete all runs without crashing.
In contrast, our method can reliably fly at high-speeds through all environments, achieving an average success rate of $70\%$ at \SI{10}{\meter\per\second}.
The drop in performance of the two baselines can be justified as follows.
Although the reactive baseline has a low processing latency, yet it has a limited expressive power: the trajectory to be executed can only be selected from a relatively small set of primitives.
This set is usually too small to account for all possible obstacle configurations and relative speeds between obstacles and drone, which are observed during high-speed flight.
In addition, being the reactive baseline only conditioned on the current observation, it is strongly affected by noise in the observation.
In contrast, the FastPlanner baseline can reject outliers in the depth map by leveraging multiple observations, which makes it more robust to sensing errors.
However, this methods generally results in higher processing latency: Multiple observations are required to add obstacles in the map, and, therefore, to plan trajectories to avoid them.
This problem is worsened by high-speed motion, which generally results in little overlap between consecutive observations.
By contrast, our approach has very the lowest processing latency (cf. section~\ref{sec:computational_cost}). In addition, the data-driven approach allows leveraging regularities in the data, which makes it more robust to sensor noise.

We thoroughly analyze those aspects in controlled studies on latency and sensor noise, which are shown in section~\ref{sec:computational_cost} and section~\ref{sec:noise_robustness}, and additionally study the robustness of our method to noise in estimation and control in section~\ref{sec:noise_sensitivity}.
A video demonstrating the performance of our method in the simulation environments can be found in \href{https://youtu.be/ALs60ij8JA8}{Movie S2}.
We now describe the specific characteristics of all environments and their respective experimental results in detail.

\subsection*{Forest}
We build a simulated forest~\cite{karaman2012high} in a rectangular region $R(l,w)$ of width $w=\SI{30}{\meter}$ and length $l=\SI{60}{\meter}$, and fill it with trees that have a diameter of about \SI{0.6}{\meter}. Trees are randomly placed according to a homogeneous Poisson point process $P$ with intensity $\delta=1/25$~tree~$\SI{}{\per\meter\squared}$~\cite{karaman2012high}.
As it is possible to see in Figure~\ref{fig:sim_results}, the resulting forest is very dense.
We provide the drone with a straight reference trajectory of \SI{40}{\meter} length, and vary the average forward speed of the drone between 3 and 10 \SI{}{\meter\per\second}.
We repeat the experiments with $10$ different random realizations of the forest, using the same random seed for all baselines.
At a low speed of \SI{3}{\meter\per\second}, all methods successfully complete every run.
As speed increases, the success rates of the baselines quickly degrade. At \SI{10}{\meter\per\second}, no baseline completes even a single run successfully.
In contrast, our approach is very robust at higher speeds. It achieves 100\% success rate 
up to \SI{5}{\meter\per\second}. At the highest speed of \SI{10}{\meter\per\second}, our approach still keeps a success rate of $60\%$.
An additional study of performance with respect to tree density $\delta$ is available in section~\ref{sec:performance_vs_density}.

\subsection*{Narrow gap} 
\begin{figure*}
\includegraphics[width=\textwidth]{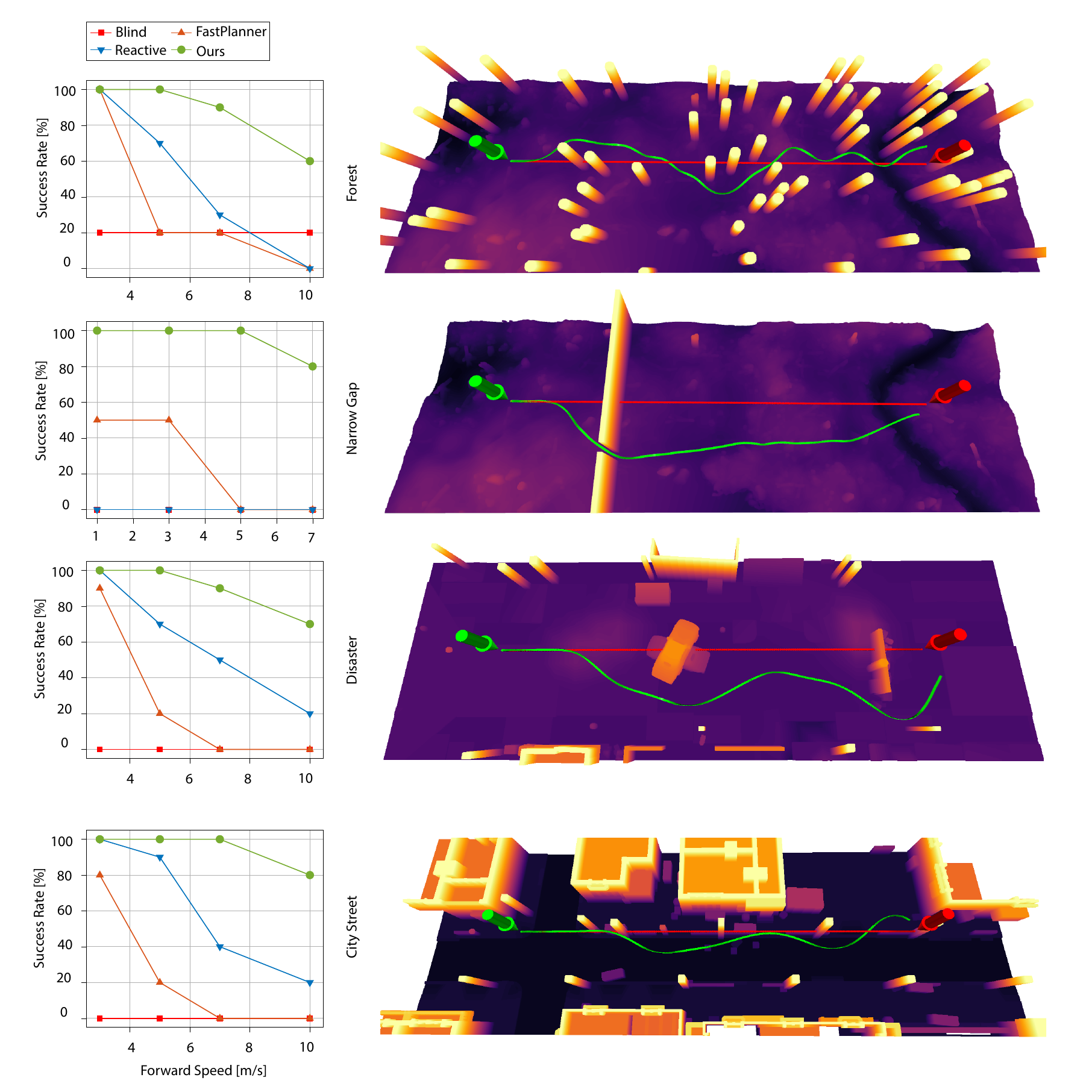}
\caption{\textbf{Experiments in a variety of simulated environments}. The left column reports success rates at various speeds. The right column shows a pointcloud of the environment together with paths taken by different policies from start (green arrow) to end (red arrow). The paths illustrate the blind policy (red) and the path taken by our approach (green). Our approach consistently outperforms the baselines in all environments and and at all speeds. Sample observations from these environments are shown in Figure~\ref{fig:sim_environments}. \label{fig:sim_results}}
 \vspace{-4ex}
\end{figure*}

We then mimic in simulation the real-world narrow gap experiment.
We render a \SI{40}{\meter} long wall with a single opening, \SI{10}{\meter} in front of the drone. The gap is placed at a randomized lateral offset in the range of $[-5,5]$\SI{}{\meter} with respect to the starting location.
The width of the opening is also randomized uniformly between \SI{0.8}{\meter} and \SI{1.0}{\meter}.
All experiments are repeated 10 times with different opening sizes and lateral gap offsets for each speed. %
An illustration of the setup and the results of the evaluation are shown in Figure~\ref{fig:sim_results}.
The reactive and blind baselines consistently fail to solve this task, while FastPlanner has a success rate of up to 50\% at \SI{5}{\meter\per\second}, but still consistently fails at \SI{7}{\meter\per\second}. 
We observe that the baselines adapt their trajectory only when being close to the wall, which is often too late to correct course toward the opening.
Conversely, our approach always completes the task successfully at speeds of up to \SI{5}{\meter\per\second}. Even at a speed of \SI{7}{\meter\per\second} it only fails in 2 of 10 runs.

\subsection*{Disaster and urban scenario} We further test our method in a simulated disaster zone and an urban city street.
These environments are challenging since they contain obstacles with complex geometry.
For instance, the former contains, among others, collapsed buildings, crashed cars, and iron fences, while the latter features street lamp poles, parked vehicles, and traffic signs.
Some observations from these environments are shown in Figure~\ref{fig:sim_environments}.
We start the drone from a random location in the map, and provide it with a straight line reference of $\SI{40}{\meter}$ in length.
For each starting location, the reference is guaranteed to be in collision with at least one obstacle.
We vary the average flight speed from \SI{3}{\meter\per\second} to \SI{10}{\meter\per\second}.
The results of these experiments as well as the pointclouds from one of the starting locations are shown in Figure~\ref{fig:sim_results}.
Because of the complex obstacle geometries, even at a low speed of $\SI{3}{\meter\per\second}$ the mapping baseline does not achieve 100\% success rate.
The reactive baseline performs better overall, with a success rate of up to 90\% at $\SI{5}{\meter\per\second}$.
However, its performance quickly degrades at higher speeds, with as much as 8 of 10 failures at \SI{10}{\meter\per\second}.
Similarly to the narrow-gap experiments, we observe that the baselines only adapt their motion when very close to obstacles, which is often too late to avoid collision in presence of large obstacles with complex shapes.
In contrast to the baselines, our approach is much more robust at higher speeds, with a success rate of up to 100\% at \SI{7}{\meter\per\second}.
Even at a speed of \SI{10}{\meter\per\second} it only fails in 2 of 10 runs in the urban environment.
The ability of our approach to combine long-term and short-term planning is crucial to achieving this
performance, since the drone must steer early enough toward free space and at the same time perform small reactive corrections to avoid a collision. 
This ability, in addition to the low latency and robustness to sensor noise, gives our approach a substantial performance advantage with respect to the baselines in these challenging environments.

\subsection{Computational cost}\label{sec:computational_cost}

\begin{table*}[t!]
\centering
\begin{tabular}{@{}c|c|cccc@{}}
\toprule
Method                       & Components                 & $\mu$ [$\SI{}{\milli \second}$] & $\sigma$ [$\SI{}{\milli \second}$] & Perc. [$\SI{}{\%}$]         & Total Proc. Latency [$\SI{}{\milli \second}$]        \\
\midrule
\multirow{3}{*}{FastPlanner~\cite{zhou2019robust}} & Pre-processing             &          14.6                &          2.3                    &             22.3                  & \multirow{3}{*}{65.2} \\
                             & Mapping                    &          49.2                &           8.7                    &            75.5                  &                    \\
                             & Planning                   &           1.4                &           1.6                    &             2.2                  &                    \\
\midrule
\multirow{2}{*}{Reactive~\cite{florence2020integrated}}    & Pre-processing            &            13.8              &            1.3                  &              72.3               & \multirow{2}{*}{19.1} \\
                             & Planning                   &             5.3              &           0.9                   &              27.7               &                    \\
\midrule 
\multirow{3}{*}{Ours}        & Pre-processing             &             0.1              &          0.04                   &            3.9                    & \multirow{3}{*}{10.3 (2.6*)} \\
                             & NN inference               &             10.1 (2.4*)      &          1.5 (0.8*)                   &            93.0                  &                    \\
                             & Projection                 &             0.08            &           0.01                  &             3.1                   &                     \\
\midrule
\multirow{3}{*}{Ours (Onboard)}        & Pre-processing             &             0.2              &          0.1                  &           0.4                    & \multirow{3}{*}{41.6} \\
                             & NN inference               &             38.9              &         4.5                &          93.6                  &                    \\
                             & Projection                 &             2.5            &           1.5                  &             6.0                  &                     \\
\bottomrule
\end{tabular}
\caption{\textbf{Processing latency} ($\mu$) on a desktop computer equipped with a hexacore i7-8700 CPU and a GeForce RTX 2080 GPU. The standard deviation $\sigma$ is computed over 1000 samples. For our approach, we report the computation time on the CPU and GPU (marked with *) on the desktop computer (\emph{Ours}), as well as the computation time on the onboard computation unit Jetson TX2, (\emph{Ours Onboard}). For the FastPlanner~\cite{zhou2019robust} and Reactive~\cite{florence2020integrated} baselines pre-processing represents the time to build a pointcloud from the depth image, while for our method pre-processing is the time to convert depth into an input tensor for the neural network. More details on the subtasks division are available in the Appendix in section~\ref{sec:computational_cost_supp}. The proposed methodology is significantly faster than prior works.
}
\label{tab:latency}
\end{table*}

In this section, we compare the computational cost of our algorithm with the baselines.
Table~\ref{tab:latency} shows the results of this evaluation. It highlights how each step of the methods contributes to the overall processing latency.
All timings were recorded on a desktop computer with a 6-core i7-8700 central processing unit (CPU), which was also used to run the simulation experiments. 
To ensure a fair comparison, we report the timings when using only the CPU for all approaches.
We also report the timings of our approach when performing neural network inference on a GeForce RTX 2080 graphics
processing unit (GPU), because 
accelerators can be used with our approach without any extra effort.
To paint a complete and realistic picture, we additionally evaluate the timing of our algorithm on the onboard computer of the quadrotor which is a Jetson TX2.

With a total computation time of \SI{65.2}{\milli\second} per frame, FastPlanner incurs the highest processing latency.
It is important to note that the temporal filtering operations that are necessary to cope with sensing errors effectively make perception even slower.
Two to three observations of an obstacle can be required to add it to the map, which increases the effective latency of the system.
By foregoing the mapping stage altogether, the \textit{Reactive} baseline significantly reduces computation time.
This baseline is about three times faster than \textit{FastPlanner}, with a total processing latency of \SI{19.1}{\milli\second}.
However, the reduced processing latency comes at the cost of the %
trajectory complexity that can be represented, since the planner can only select primitives from a pre-defined library.
In addition, the reactive baseline is sensitive to sensing errors, which can drastically affect performance at high speeds.

Our approach has significantly lower processing latency than both baselines; when network inference is performed on the GPU, our approach is 25.3 times faster than \textit{FastPlanner} and 7.4 times faster than the \textit{Reactive} baseline.
When GPU inference is disabled, the network's latency increases by only \SI{8}{\milli\second}, and our approach is still much faster than both baselines.
Moving from the desktop computer to the onboard embedded computing device, the network's forward pass requires \SI{38.9}{\milli\second}.
Onboard, the total time to pass from the sensor reading to a plan is \SI{41.6}{\milli\second}, which corresponds to an update rate of about \SI{24}{\hertz}.

\subsection{The effect of latency and sensor noise}\label{sec:noise_robustness}
\begin{figure*}
\centering
\includegraphics[width=\linewidth]{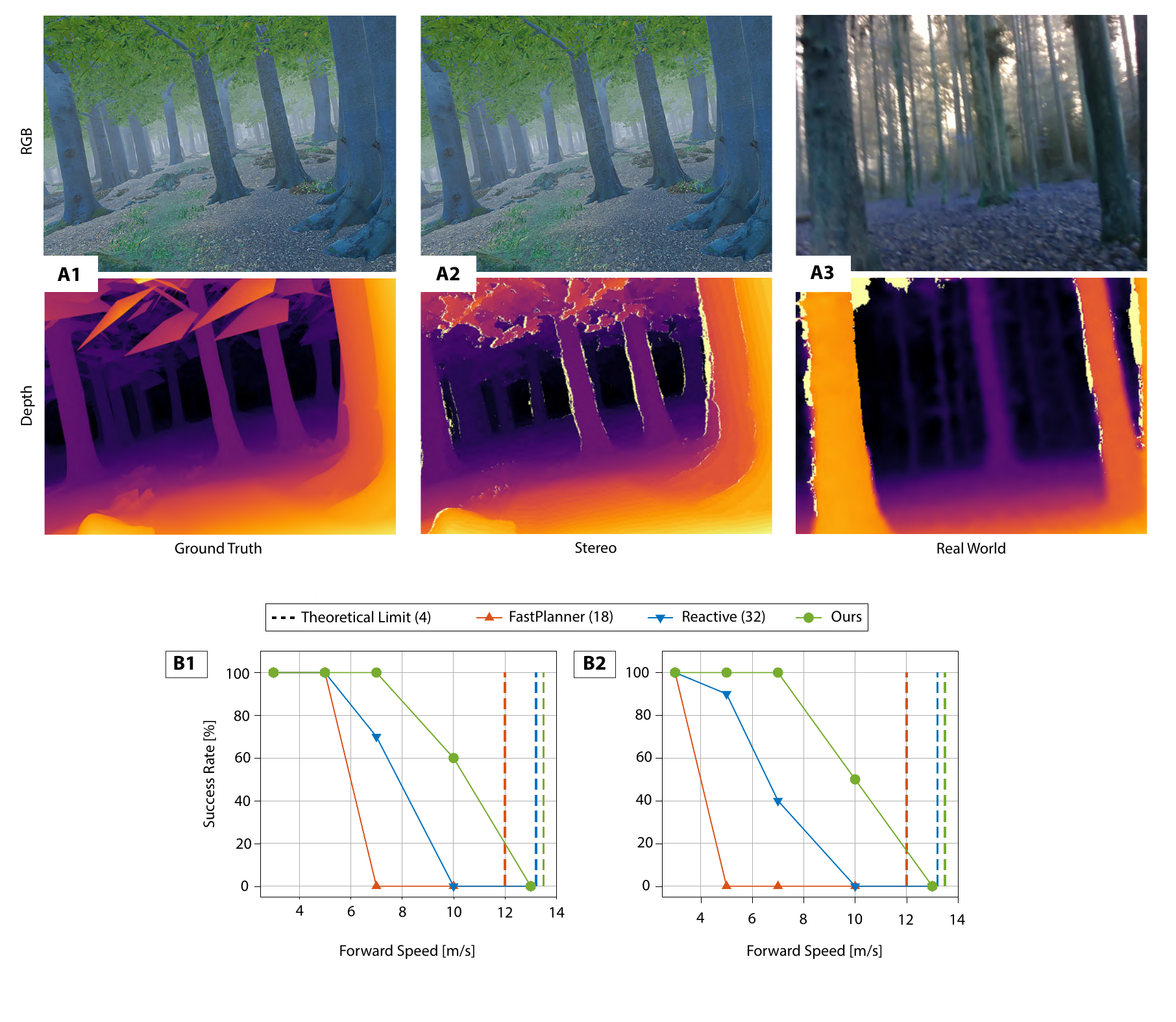}
\caption{\textbf{The effect of sensor noise on performance}. (\textbf{A}) When comparing RGB and depth images generated in simulation with images captured in the real world, we observe that the corresponding depth images are more similar than the RGB images. In addition, simulated depth estimated by stereo matching (\textbf{A2}) contains the typical failure cases of a depth sensor (\textbf{A3}), \emph{e.g.} missing values and noise. (\textbf{B}) Results of the controlled experiment to study the effect of perception latency and noise on navigation ability. The experiment is performed on ground-truth depth (\textbf{B1}) and stereo depth (\textbf{B2}). Since each method has different processing latency (c.f. section~\ref{sec:computational_cost}), the theoretical maximum speed differs between approaches (dashed lines). Our method can fly closer to the theoretical limit than the baselines and is only minimally affected by the noise of stereo depth estimates.} \label{fig:result_pole}
\end{figure*}

We analyze the effect of sensor noise and planning latency in a controlled experiment.
In this experiment, the quadrotor travels along a straight line at a constant forward speed and is required to laterally evade a single obstacle (a pole) while having only limited sensing range.
This experimental setup was proposed in Falanga et al.~\cite{Falanga19ral_howfast} to understand the role of perception latency on the navigation ability of a robotic system subject to bounded inputs.
Specifically, the authors derived an upper-bound for the forward speed at which a robot can fly and avoid a single obstacle as a function of perception latency and sensing range.
They modeled the robot as a point-mass, which is a limited approximation for a quadrotor as it neglects the platform's rotational dynamics.
We thus extend their formulation to account for the latency introduced by the rotational motion that is necessary to avoid the obstacle.
A detailed description of the formulation can be found in the Supplementary Materials in section~\ref{sec:rotation_effects}.

We set up the experiment by placing a quadrotor with an initial forward velocity $v$ at a distance of $\SI{6}{\meter}$ from a pole with diameter $\SI{1.5}{\meter}$.
The quadrotor is modeled as a sphere with radius $\SI{0.2}{\meter}$.
According to our formulation, we compute a theoretical maximum speed---\emph{i.e.} the speed at which the task is no longer feasible---for each method.
The maximum speed depends on the sensing range, \emph{i.e.} how far can an obstacle be accurately perceived, the latency of the visual sensor, \emph{i.e.} the inverse of the frame rate, and the processing latency, \emph{i.e.} the time to convert an observation into motor commands.
Note that the latter varies between methods (c.f. section~\ref{sec:computational_cost}), and therefore the theoretical maximum speed differs for each approach.
For the details of the calculation, we refer the reader to section~\ref{sec:rotation_effects} in the Supplementary Materials.
We then perform the controlled experiment with varying forward speeds $v$ in the range of 3 to 13 $\SI{}{\meter\per\second}$.
We perform 10 experiments for each speed with all approaches and report the success rate. 
We run the experiment in two settings: (i) with ground-truth depth information, to isolate the effect of latency on performance, and (ii) with depth estimated by stereo matching~\cite{hirschmuller2007stereo} to analyze the effect of sensing errors to performance.

\subsection*{Ground-truth depth}
Figure~\ref{fig:result_pole}B1 illustrates the results of this experiment when perfect depth perception (Figure~\ref{fig:result_pole}A1) is available.
All approaches can complete the task perfectly up to 5$\SI{}{\meter\per\second}$.
However, even in these ideal conditions, the performance of the baselines drops for speeds beyond 5$\SI{}{\meter\per\second}$.
This drop in performance for \textit{Reactive} can be attributed to the fact that the finite library of motion primitives does not contain maneuvers that are aggressive enough to complete this task.
Similarly, the performance degrades for \textit{FastPlanner} as a result of sub-optimal planning of actions.
Although this baseline manages to map the obstacle in time, the planner frequently commands actions to stop the platform which leads to crashes when flying at high speeds.
It is important to point out that the \textit{FastPlanner} baseline was only demonstrated up to speeds of  3$\SI{}{\meter\per\second}$ in the original work~\cite{zhou2019robust}, and thus was not designed to operate at high speeds.
Our approach can successfully avoid the obstacle without a single failure up to $\SI{7}{\meter\per\second}$.
For higher speeds, performance gracefully degrades to 60\% at $\SI{10}{\meter\per\second}$.
This decrease in performance can be attributed to the sensitivity to imperfect network predictions when flying at high speed, where
a single wrong action can lead to a crash.

\subsection*{Estimated depth}
While the previous experiments mainly focused on latency and the avoidance strategy, we now study the influence of imperfect sensory measurements on performance.
We repeat the same experiment, but provide all methods with depth maps that have been computed from the stereo pairs (Figure~\ref{fig:result_pole}A2).
Figure~\ref{fig:result_pole}B2 shows the results of this experiment.
The baselines experience a significant drop in performance compared with when provided with perfect sensory readings.
\textit{FastPlanner} completely fails for speeds of $\SI{5}{\meter\per\second}$ and beyond.
This sharp drop in performance is because of the need for additional filtering of the noisy depth measurements that drastically increases the latency of the mapping stage. 
To reject outliers from the depth sensor, multiple observation (two/three) are required to completely map the obstacle.
As a result, this baseline detects obstacles too late to be able to successfully evade them.
Similarly, the performance of the \textit{Reactive} baseline drops by 30\% at $\SI{7}{\meter\per\second}$.
In contrast to the baselines, our approach is only marginally affected by the noisy depth readings, with only a 10\% drop in performance at \SI{10}{\meter\per\second}, but no change in performance at lower speeds.
This is because our policy, trained on depth from stereo, learns to account for common issues in the data such as discretization artifacts and missing values.

\section{Discussion} \label{sec:discussion}

Existing autonomous flight systems are highly engineered and modular. The navigation task is usually split into sensing, mapping, planning, and control. The separation into multiple modules simplifies the implementation of engineered systems, enables parallel development of each component, and makes the overall system more interpretable. However, modularity comes at a high cost: the communication between modules introduces latency, errors compound across modules, and interactions between modules are not modeled. In addition, it is an open question if
certain subtasks, such as maintaining an explicit map of the environment, are even necessary for agile flight.

Our work replaces the traditional components of sensing, mapping, and planning with a single function that is represented by a neural network.
This increases the system's robustness against sensor noise and reduces the processing latency.
We demonstrate that our approach can reach speeds of up to \SI{10}{\meter\per\second} in complex environments and reduces
the failure rate at high speeds by up to 10 times when compared with the state of the art.

We achieve this by training a neural network to imitate an expert with privileged information in simulation.
To cope with the complexity of the task and to enable seamless transfer from simulation to reality, we make several technical contributions. These include a sampling-based expert, a neural network architecture, and a training procedure, all of which take the task's multi-modality into account. We also use an abstract, but sufficiently rich input representation that considers real-world sensor noise.
The combination of these innovations enables the training of robust navigation policies in simulation that can be directly transferred to diverse real-world environments without any fine-tuning on real data.

We see several opportunities for future work.
Currently, the learned policy exhibits low success rates  at average speeds of \SI{10}{\meter\per\second} or higher in the real world.
At these speeds even our expert policy, despite having perfect knowledge of the environment, often fails to find collision-free trajectories.
This is mainly because of the fact that, at speeds of \SI{10}{\meter\per\second} or higher, feasible solutions require temporal consistency over a long time horizon and strong variations of the instantaneous flying speed as a function of the obstacle density.
This requirement makes feasible trajectories extremely sparse in parameter space, resulting in intractable sampling.
Engineering a more complex expert to tackle this problem can be very challenging and might require specifically tailored heuristics to find approximate solutions.
Therefore, we believe that this problem represents a big opportunity for model-free methods, which have the potential to ease the engineering requirements.
The second reason for the performance drop at very high speeds is the mismatch between the simulated and physical drone in terms of dynamics and perception.
The mismatch in dynamics is because of aerodynamics effects, motor delays, and dropping battery voltage.
Therefore, we hypothesize that performance would benefit from increasing the fidelity of the simulated drone and making the policy robust to the unavoidable model mismatches.
A third reason is perception latency: Faster sensors can provide more information about the environment in a smaller amount of time and, therefore, can be used to provide more frequent updates, which improves the precision of the estimate.
This could enable further reducing sensitivity to noise and promote a quicker understanding of the environment.
Perception latency can possibly be reduced with event cameras~\cite{Gallego20pami}, especially in the presence of dynamic obstacles~\cite{Falanga20SR}.

Overall, our approach is a stepping stone toward the development of autonomous systems that can navigate at high speeds through previously unseen environments with only on-board sensing and computation.
Combining our short-horizon controller for local obstacle avoidance with a long-term planner is a major opportunity for many robotics applications, including autonomous exploration, delivery, and cinematography.

\section{Materials and Methods}

\begin{figure*}[t]
    \centering
    \includegraphics[width=\textwidth]{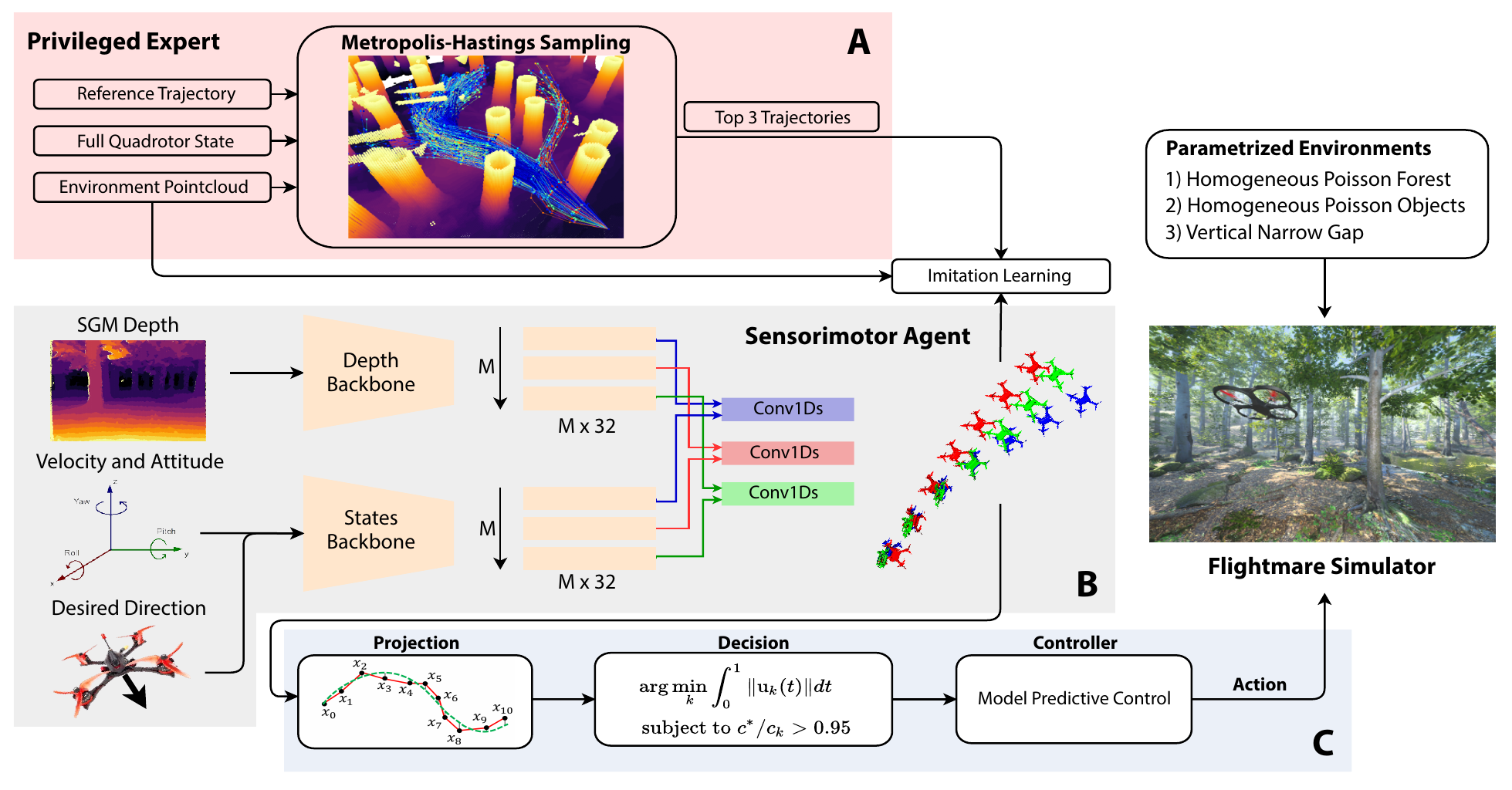}
    \caption{ \textbf{Method overview}. (\textbf{A}) Our offline planning algorithm computes a distribution of collision-free trajectories to follow a reference trajectory. The trajectories are computed with Metropolis-Hastings sampling and are conditioned on complete 3D knowledge of the environment, which is represented by a point cloud. (\textbf{B}) A sensorimotor agent is trained with imitation learning to predict the best three trajectories from the estimated depth, the drone's velocity and attitude, and the desired direction that encodes the goal. (\textbf{C}) The predictions are projected on the space of polynomial trajectories and ranked according to their predicted collision cost $c_k$. 
    The trajectory with the lowest predicted cost $c_k$ is then tracked with a model predictive controller. 
    If multiple trajectories have similar predicted cost (within a $5\%$ range of the minimum $c^*=\min c_k$), the one with the smallest actuation cost is used. 
    }
    \label{fig:method_overview}
\end{figure*}

To perform agile flight through cluttered and previously-unseen environments, we train a sensorimotor policy to predict receding-horizon trajectories from on-board sensor measurements and a reference trajectory.
We assume that the reference trajectory is provided by a higher-level planning algorithm or the user.
The reference is not necessarily collision-free and only encodes the long-term goal of the platform.
The agent is responsible to fly in the direction dictated by the reference while adapting its flight path to the environment.

An agent's observation $\bm{o}$ consists of a depth image $\bm{d} \in \real^{640\times480}$, an estimate of the platform velocity $\bm{v} \in \real^{3}$  and attitude (expressed as a rotation matrix) $\bm{q} \in \real^{9}$, and a desired flight direction $\bm{\omega} \in \real^{3}$.
The policy output is a set of motion hypotheses which are represented as receding-horizon trajectories with the corresponding estimated risk of collision. 
A model predictive controller then tracks the trajectory with the lowest collision probability and input cost.
The controller is trained using privileged learning~\cite{chen2020learning}.
Specifically, the policy is trained on demonstrations provided by a sampling-based planner that has access to information that is not available to the sensorimotor student: the precise knowledge of the 3D structure of the environment and the exact platform state.
Figure~\ref{fig:method_overview} shows an overview of our method.

We collect demonstrations and perform training entirely in simulation. This trivially facilitates access to perfect 3D and state data for the privileged expert, enables the synthesis of unlimited training samples for any desired trajectory, and does not put the physical platform in danger.
We use the Flightmare simulator~\cite{yunlong2020flightmare} with the RotorS~\cite{furrer2016rotors} Gazebo plugin and Unity as a rendering engine~\cite{juliani2018unity}.
Both the training and the testing environments are built by adding obstacles to the uneven ground of an \href{https://assetstore.unity.com/packages/3d/vegetation/forest-environment-dynamic-nature-150668}{out-of-the-box Unity environment}.

To enable zero-shot transfer to real environments, special care has to be taken to minimize the domain shift of the 
(visual) input modalities from simulation to reality. To this end we use depth as an abstract input representation that
shows only a negligible domain shift from simulation to the real world (cf. Figure~\ref{fig:result_pole}A).
Specifically, the sensorimotor agent is trained with depth images that have been computed using Semi-Global Matching (SGM)~\cite{hirschmuller2007stereo} from a simulated stereo camera pair.
As off-the-shelf depth sensors, such as the Intel RealSense~435 camera used on our physical platform, compute depth from stereo images using similar principles~\cite{Keselman_2017_CVPR_Workshops}, this strategy ensures that the characteristics of the input aligns between simulation and the real world.
As a result, the trained sensorimotor agent can be directly deployed in the real world. 
The next section presents both the privileged expert and the sensorimotor agent in detail.

\subsection{The privileged expert} \label{sec:expert_planner}
Our privileged expert is a sampling-based motion planning algorithm.
The expert has perfect knowledge of the platform state and the environment (complete 3D map), both of which are only available in simulation.
The expert generates a set of collision-free trajectories $\bm{\tau}$ %
representing the desired state of the quadrotor $\bm{x}_{des} \in \real^{13}$ over the next second, starting from the current state of the drone, i.e. $\bm{\tau}(0) = \bm{x}$.
To do so, it samples from a probability distribution $P$ that encodes distance from obstacles and proximity to the reference trajectory.
Specifically, the distribution of collision-free trajectories $P(\bm{\tau}\;|\;\bm{\tau}_{\text{ref}}, \mathcal{C})$ is conditioned on 
the reference trajectory $\bm{\tau}_{\text{ref}}$ and the structure of the environment in the form of a point cloud $\mathcal{C} \in \real^{n\times3}$.
According to $P$, the probability of a trajectory $\bm{\tau}$ is large if far from obstacles and close to the reference $\bm{\tau}_{\text{ref}}$.
We define $P$ as the following:
\begin{align}
    P(\bm{\tau}\;|\;\bm{\tau}_{\text{ref}}, \mathcal{C}) &= \frac{1}{Z} \exp(-c(\bm{\tau}, \bm{\tau}_{\text{ref}}, \mathcal{C}))
\end{align}
where $Z=\int_{\bm{\tau}} P(\bm{\tau}\;|\; \bm{\tau}_{\text{ref}}, \mathcal{C})$ is the normalization factor and  $c(\bm{\tau}, \bm{\tau}_{\text{ref}}, \mathcal{C}) \in \real_{+}$ is a cost function indicating proximity to the reference and distance from obstacles.
We define the trajectory cost function as
\begin{align}\label{eq:traj_cost}
    c(\bm{\tau}, \bm{\tau}_{\text{ref}}, \mathcal{C}) = &\int_{0}^{1} \lambda_c C_{collision}(\bm{\tau}(t)) + \notag \\
    &\int_{0}^{1}
    [\bm{\tau}(t) - \bm{\tau}_{ref}(t)]^\top \bm{Q}[\bm{\tau}(t) - \bm{\tau}_{ref}(t)] \; dt
\end{align}
where $\lambda_c=1000$, $\bm{Q}$ is a positive semidefinite state cost matrix, and $C_{collision}$ is a measure of the distance of the quadrotor to the points in $\mathcal{C}$.
We model the quadrotor as a sphere of radius $r_q=\SI{0.2}{\meter}$ and define the collision cost as a truncated quadratic function of $d_c$, \emph{i.e.} the distance between the quadrotor and the closest point in the environment:
\begin{equation}~\label{eq:collision_cost}
C_{collision}(\bm{\tau}(t)) = \begin{cases}
0 &\text{if } d_c > 2 r_q \\
- d_c^2 / r_q^2 + 4 &\text{otherwise}.
\end{cases}
\end{equation}
The distribution $P$ is complex because of the presence of arbitrary obstacles and frequently multi-modal in cluttered environments since obstacles can be avoided in multiple ways.
Therefore, the analytical computation of $P$ is generally intractable.

To approximate the density $P$, the expert uses random sampling.
We generate samples with the M-H algorithm~\cite{hastings70} as it provides asymptotic convergence guarantees to the true distribution.
To estimate $P$, the M-H algorithm requires a target score function $s(\bm{\tau}) \propto P(\bm{\tau}\;|\;\bm{\tau}_{\text{ref}}, \mathcal{C})$.
We define $s(\bm{\tau}) = \exp (-c(\bm{\tau}, \bm{\tau}_{\text{ref}}, \mathcal{C}))$, where $c(\cdot)$ is the cost of the trajectory $\bm{\tau}$.
It is easy to show that this definition satisfies the conditions for the M-H algorithm to asymptotically estimate the target distribution $P$.
Hence, the trajectories sampled with M-H will asymptotically cover all of the different modes of $P$.
We point the interested reader to the Supplementary Materials (section~\ref{sec:metropolis}), for an overview of the M-H algorithm and its convergence criteria.

To decrease the dimension of the sampling space, we use a compact yet expressive representation of the trajectories $\bm{\tau}$.
We represent $\bm{\tau}$ as a cubic B-spline $\bm{\tau}_{bspline} \in \real^{3\times3}$ curve with three control points and a uniform knot vector, enabling interpolation with high computational efficiency~\cite{Gallier99bspline}.
Cubic B-splines are twice continuously differentiable and have a bounded derivative in a closed interval.
Because of the differential flatness property of quadrotors~\cite{mellinger2011minimum}, continuous and bounded acceleration directly translates to continuous attitude over the trajectory duration.
This encourages dynamically feasible trajectories that can be tracked by a model-predictive controller accurately ~\cite{mellinger2011minimum, falanga2018pampc}.
Therefore, instead of naively sampling the states of $\bm{\tau}$, we vary the shape of the trajectory by sampling the control points of the B-spline in a spherical coordinate system.
In addition, for computational reasons, we discretize the trajectory at equally spaced time intervals of \SI{0.1}{\second} and evaluate the discrete version of Equation~\ref{eq:traj_cost}.
Specifically, we sample a total of $50$ thousands trajectories using a Gaussian with a variance of ${2,5,10}$ that increases every $16$ thousands samples as the proposal distribution. 
Despite this efficient representation, the privileged expert cannot run in real-time, given the large computational overhead introduced by sampling.

To bias the sampled trajectories toward obstacle-free regions, we replace the raw reference trajectory $\bm{\tau}_{ref}$ in  Equation~\ref{eq:traj_cost} with a global collision-free trajectory $\bm{\tau}_{gbl}$ from start to goal, that we compute using the approach of Liu et al.~\cite{liu2018search}.
As illustrated in Figure~\ref{fig:global_plan_helps}, conditioning sampling on $\bm{\tau}_{gbl}$ practically increases the horizon of the expert and generates more conservative trajectories.
An animation explaining our expert is available in \href{https://www.youtube.com/watch?v=m89bNn6RFoQ}{Movie 1}.
After removing all generated trajectories in collision with obstacles, we select the three best trajectories with lower costs.
Those trajectories are used to train the student policy.

\subsection{The student policy} \label{sec:learner}

In contrast to the privileged expert, the student policy produces collision-free trajectories in real time with access only to on-board sensor measurements.
These measurements include a depth image estimated with SGM~\cite{hirschmuller2007stereo}, the platform velocity and attitude, and the desired direction of flight.
The latter is represented as a normalized vector heading toward the reference point 1 s in the future with respect to the closest reference state.
We hypothesize that this information is sufficient for generating the highest probability samples of the distribution $P(\bm{\tau}\;|\; \bm{\tau}_{\text{ref}}, \mathcal{C})$ without actually having access to the point cloud $\mathcal{C}$.
There are two main challenges to accomplishing this: (i) The environment is only partially observable from noisy sensor observations, and (ii) the distribution $P$ is in general multi-modal. Multiple motion hypotheses with high probabilities can be available, but their average can have a very low probability.

We represent the policy as a neural network that is designed to be able to cope with these issues.
The network consists of an architecture with two branches that produce a latent encoding of visual, inertial, and reference information, and outputs $M=3$ trajectories and their respective collision cost.
We use a pre-trained MobileNet-V3 architecture~\cite{HowardICCV19} to efficiently extract features from the depth image.
The features are then processed by a 1D convolution to generate $M$ feature vectors of size $32$.
The current platform's velocity and attitude are then concatenated with the desired reference direction and processed by a four-layer perceptron with $[64,32,32,32]$ hidden nodes and LeakyReLU activations. We again use 1D convolutions to create a $32$-dimensional feature vector for each mode.
The visual and state features are then concatenated and processed independently for each mode by another four-layer perceptron with $[64,128,128]$ hidden nodes and LeakyReLU activations.
The latter predicts, for each mode, a trajectory $\bm{\tau}$ and its collision cost.
In summary, our architecture receives as input a depth image $\bm{d} \in \real^{640\times480}$, the platform's velocity $\bm{v} \in \real^3$, the drone's attitude expressed as a rotation matrix $\bm{q} \in \real^9$, and reference direction $\bm{\omega} \in \real^{3}$.
From this input it predicts a set $\mathcal{T}_n$ of trajectories and their relative collision cost, \ie $\mathcal{T}_n = \lbrace (\bm{\tau}_n^k, c_k) \; \vert \; k \in [0, 1,\ldots, M-1]  \rbrace$, where $c_k \in \real_+$. 
Differently from the privileged expert, the trajectory predicted by the network does not describe the full state evolution but only its position component, i.e. $\bm{\tau}_n^k \in \real^{10\times3}$.
Specifically, the network trajectory $\bm{\tau}_n^k$ is described by:
\begin{equation}
    \bm{\tau}_n^k = \left[ \; \bm{p}(t_i) \; \right]_{i=1}^{10}, \quad t_i = \frac{i}{10}, 
\end{equation}
where $\bm{p}(t_i) \in \real^3$ is the drone's position at time $t=t_i$ relative to its current state $\bm{x}$.
This representation is more general than the B-spline with three control points used by the sampling-based planner, and it was preferred to the latter representation to avoid the computational costs of interpolation at test time.  %

\begin{figure*}[t]
\includegraphics[width=\textwidth]{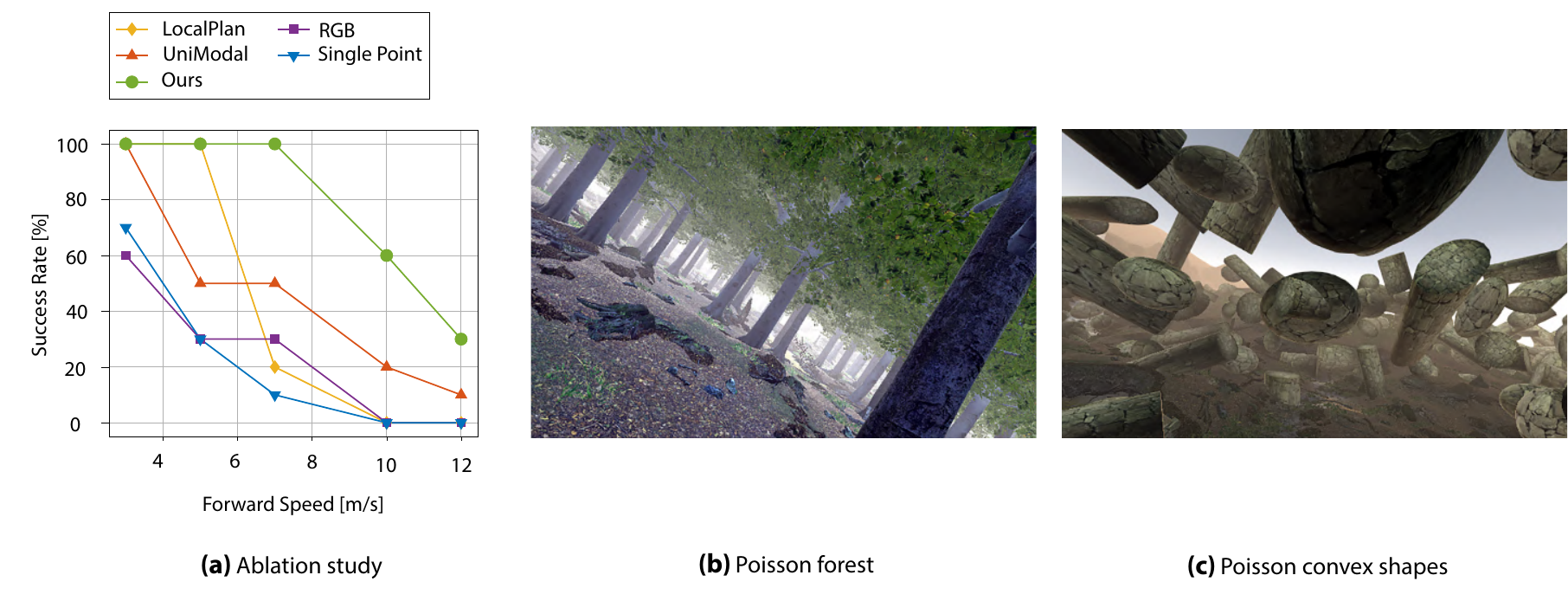}
\caption{(a) \textbf{Method validation}. With no initialization of the sampler on a global trajectory (\emph{LocalPlan}), no multi-modal training (\emph{UniModal}), or no training on SGM depth image (\emph{RGB}), performance significantly drops. Predicting a single point in the future instead of a full trajectory (\emph{SinglePoint}) has similar effects on performance.  \emph{Ours} consistently outperforms all the ablated versions of the system. (b-c) Training environments. We build training environments by spawning trees or convex shapes according to an homogeneous Poisson point process with varying intensities. Additional samples of the training and testing simulation environments are available in Figure~\ref{fig:sim_environments}. \label{fig:ablation_studies}}
\end{figure*}

We train the neural network with supervised learning on the three trajectories with lowest cost found by the expert.
To account for the multi-hypotheses prediction, we minimize the following \emph{Relaxed Winner-Takes-All} (R-WTA) loss for each sample
\begin{equation}
    \text{R-WTA}(\mathcal{T}_e, \mathcal{T}_n) = \sum_{i=0}^{|\mathcal{T}_e|} \sum_{k=0}^{|\mathcal{T}_n|} \alpha(\traj_{e,p}^i,\traj_n^k) \| \traj_{e,p}^i - \traj_n^k \|^2,
    \label{eq:minimum_voronoi}
\end{equation}
where $\mathcal{T}_e$ and $\mathcal{T}_n$ are the set of expert and network trajectories, $\traj_{e,p}$ denotes the position component of $\traj_{e}$, and $\alpha(\cdot)$ is defined as
\begin{equation}
\alpha(\traj_{e, p}^i,\traj_n^k) = \begin{cases}
1-\epsilon &\text{if  $\| \traj_{e, p}^i - \traj_n^k \|^2 \leq  \| \traj_{e, p}^j - \traj_n^k \|^2 \quad \forall j \neq i$} \\
\frac{\epsilon}{M-1} &\text{otherwise}.
\end{cases}
\label{eq:epsilon}
\end{equation}
A necessary condition for Equation~\ref{eq:minimum_voronoi} to be minimized is that the network predictions $\mathcal{T}_n$ correspond to the centroids of a Voronoi tessellation of expert labels $\mathcal{T}_e$~\cite{RupprechtICCV17}. However, Equation~\ref{eq:minimum_voronoi} is not differentiable, since it performs a hard assignment between predictions and labels. Therefore, we relax the assignment with a small value $\epsilon$ (Equation~\ref{eq:epsilon}) to optimize Equation~\ref{eq:minimum_voronoi} with gradient descent.
Intuitively, an expert trajectory $\traj_e^i \in \mathcal{T}_e$ is associated with the closest network trajectory $\traj_n^k~\in~\mathcal{T}_n$ with a weight of $1-\epsilon=0.95$ and with $\epsilon/(M-1)=0.025$ to all remaining hypotheses. 
This formulation, proposed by Rupprecht et al.~\cite{RupprechtICCV17}, prevents mode collapse and was shown to outperform other popular approaches for multi-modal learning such as Mixture Density Networks~\cite{bishop1994mixture}.
In addition, the predicted collision cost $c_k$ is trained with supervised learning on the ground-truth cost $C_{collision}(\traj_n^k)$ computed according to Equation~\ref{eq:collision_cost}.
In summary, the final training loss for each sample is equal to:
\begin{equation}
    \mathcal{L}((\mathcal{T}_e, \mathcal{T}_n) =  \lambda_1\text{R-WTA}(\mathcal{T}_e, \mathcal{T}_n) + \lambda_2\sum_{k=0}^{|\mathcal{T}_n|} \| c_k -  C_{collision}(\traj_n^k) \| ^2,
\end{equation}
where $\lambda_1=10$ and $\lambda_2=0.1$ were empirically found to equalize the magnitudes of the two terms.
This loss is averaged over a minibatch of $8$ samples and minimized with the Adam optimizer~\cite{KingmaB14} and a learning rate of $1\times10^{-3}$.

At test time we retrieve the full state information from the predicted trajectories by projecting them on the space of order-5 polynomials for each axis independently.
This representation enforces continuity in position, velocity, and acceleration with respect to the current state, and facilitates dynamic feasibility because of differential flatness~\cite{mellinger2011minimum}.
Considering for example the $x$ axis, we define the polynomial projection $\mu_x(t) = \bm{a}_x \trans \cdot \timevec(t)$, where $\bm{a}_x \trans= \left[ a_0, a_1, \ldots, a_5 \right]$ and $\timevec(t) \trans = \left[ 1, t, \ldots, t^5 \right]$.
The projection term $\bm{a}_x$ is found by solving the following optimization problem:
\begin{alignat}{2}
& \underset{\bm{a}_x}{\minimize}\quad &&
\sum_{i=1}^{10} \left(\traj_{n,x}^{k,i} - \bm{a}_x \trans \cdot \timevec(\frac{i}{10})\right)^2 \notag \\
& \text{subject to}\quad &&
s_x(0) - \bm{a}_x \trans \cdot \timevec(0) = 0 \notag \\
& && \dot{s}_x(0) - \bm{a}_x \trans \cdot \dot{\timevec}(0) = 0 \notag \\
& && \ddot{s}_x(0) - \bm{a}_x \trans \cdot \ddot{\timevec}(0) = 0
\label{eq:projection}
\end{alignat}
where $\traj_{n,x}^{k,i}$ is the $x$ component of the $i^{\text{th}}$ element of $\traj_n^k$ and $s_x(0), \dot{s}_x(0), \ddot{s}_x(0)$ are the x position of the quadrotor and its derivatives obtained from the current state estimate.
The latter corresponds to the ground-truth state when deployed in simulation and to the state estimate computed by the Intel RealSense~T265 when deployed on the physical platform.
To reduce abrupt changes in velocity during flight, which would lead to strong pitching motion, we additionally constrain the polynomial's average speed to a desired value $v_{des}$.
To do so, we scale the time $t$ of polynomial $\mu_x(t)$ by a factor $\beta = v_{des} / v_{\mu}^x$, \ie $t' = \beta t$, where $v_{\mu}^x = \| \mu(1) - \mu(0) \|$.

Once all predicted trajectories are projected we select one for execution.
To do so, we select trajectories with $c^*/c_k \geq 0.95$ ($c^*=\min c_k$) and compute their input costs according to Mellinger and Kumar~\cite{mellinger2011minimum}.
The one with the lowest input costs is tracked by a model-predictive controller~\cite{falanga2018pampc}.
Intuitively, this choice enforces temporal continuity in the trajectories.
For example, when dodging an obstacle to the right, we do not want to steer toward the left at the next iteration, unless strictly necessary because of the appearance of an obstacle.

\subsection{Training environments} \label{sec:train_env}

We build custom environments in the Flightmare simulator~\cite{yunlong2020flightmare} to collect training data.
All environments are built by spawning items on the uneven empty ground of an off-the-shelf Unity environment.
We spawn items belonging to two categories: simulated trees, available off-the-shelf, and a set of convex shapes such as ellipsoids, cuboids, and cylinders.
Sample observations from these environments are available in Figure~\ref{fig:ablation_studies}. Additional training samples, together with observations collected in the simulated testing environments, are available in Figure~\ref{fig:sim_environments}.
The dimensions of these shapes are randomized according to a continuous uniform random distribution with $x \in \mathcal{U}(0.5,4)$, $y \in \mathcal{U}(0.5,4)$, and  $z \in \mathcal{U}(0.5,8)$.
Training environments are created by spawning either of the two categories of items according to a homogeneous Poisson point process with intensity $\delta$.

We generate a total of $850$ environments by uniform randomization of the following two quantities: item category, \ie, trees or shapes, and the intensity $\delta \in \mathcal{U}(4,7)$, with $\delta \in \nat_+$.
For each environment, we compute a global collision-free trajectory $\bm{\tau}_{gbl}$ from the starting location to a point \SI{40}{\meter} in front of it.
The trajectory $\bm{\tau}_{gbl}$ is not observed by the student policy, but only by the expert.
The student is only provided with a straight, potentially not collision-free, trajectory from start to end to convey 
the goal.

To assure sufficient coverage of the state space, we use the dataset aggregation strategy (DAgger)~\cite{ross2011reduction}.
This process consists of rolling out the student policy and labeling the visited states with the expert policy.
To avoid divergence from $\bm{\tau}_{gbl}$ and prevent crashes in the early stages of training, we track a trajectory predicted by the student only if the drone's distance from the closest point on $\bm{\tau}_{gbl}$ is smaller than a threshold $\xi$, initialized to zero.
Otherwise, we directly track $\bm{\tau}_{gbl}$ with a model-predictive controller. 
Every $30$ environments the student is re-trained on all available data and the threshold $\xi$ set to $\xi'=\min(\xi+0.25, 6)$.
Aggregating data over all environments results in a dataset of about $90$K samples.
For the narrow-gap experiments, we finetune the student policy on $100$ environments created by adding a \SI{50}{\meter} long wall with a single vertical gap of random width $w_g \in \mathcal{U}(0.7,1.2)$ meters in the center.
In those environments, the drone starts at a \SI{10}{\meter} distance from the wall with a randomized lateral offset $l \in \mathcal{U}(-5,5)$ from the gap.
We collect about $10$K training samples from those environments.

We evaluate the trained policies in simulation on environments not seen during training but coming from the same distributions.
The same policies are then used to control the physical platforms in real-world environments.
When deployed in simulation, we estimate depth with SGM~\cite{hirschmuller2007stereo} from a simulated stereo pair and use the ground-truth state of the platform.
Conversely, on the physical platform depth is estimated by an off-the-shelf Intel RealSense~435 and state estimation is performed by an Intel RealSense~T265.
More details on our experimental platform are available in section~\ref{sec:experimental_platform}.

\subsection{Method validation}\label{sec:ablation_studies}

Our approach is based on several design choices that we validate in an ablation study. 
We ablate the following components: (i) the use of global planning to initialize the sampling of the privileged expert, (ii) the use of depth as an intermediate representation for action, and (iii) multi-modal network prediction.
The results in Figure~\ref{fig:ablation_studies} show that all components are important and that some choices have a larger impact than others.
Our study indicates that depth perception plays a fundamental rule for high-speed obstacle avoidance: When training on color images (\emph{RGB}) performance drops significantly.
This is inline with previous findings~\cite{Zhou19ScR, kaufmann2020RSS} showing that intermediate image representations have a strong positive effect on performance when sufficiently informative for the task.
Not accounting for multi-modal trajectory prediction (\emph{UniModal}) is also detrimental for performance. This is because of the ambiguities inherent of the task of obstacle avoidance: the average of expert trajectories is often in collision with obstacles.
Not initializing the sampling of expert trajectories on a global plan (\emph{LocalPlan}) is not important for low-speed flight, but plays an important role for success at higher speeds.
Biasing the motion away from regions that are densely populated by obstacles is particularly important for high-speed flight where little slack is available for reacting to unexpected obstacles.
In addition, we compare our output representation, \emph{i.e.} a trajectory 1 s in the future, to the less complex output representation
in prior work~\cite{Loquercio19TRO}, \emph{i.e.} a single waypoint in the future (\emph{SinglePoint}).
The results indicate that the limited representation power of the latter representation causes performance to drop significantly, especially at high speeds.

\section{Acknowledgments}
 The authors thank T. L\"angle, M. Sutter, Y. Song, C. Pfeiffer, L. Bauersfeld, and N. Aepli for their contributions to the drone design, the simulation environment, and preparation of multimedia material. We also thank J. Lee for the template of this document. \textbf{Funding}: This work was supported by the Intel Network on Intelligent Systems, Armasuisse, the National Centre of Competence in Research (NCCR) Robotics through the Swiss National Science Foundation (SNSF) and by the European Research Council (ERC) under Grant Agreement 864042 (AGILEFLIGHT). \textbf{Author contributions}: A.L and E.K. formulated the main ideas, implemented the system, performed all experiments, and wrote the paper; R.R and M.M. contributed to the experimental design, data analysis, and paper writing;  V.K. and D.S. provided funding and contributed to the design and analysis of experiments. \textbf{Competing interests}: The authors declare that they have no competing interests.\textbf{ Data and materials availability}: All (other) data needed to evaluate the conclusions in the paper are present in the paper or the Supplementary Materials. Other material can be found at \url{http://rpg.ifi.uzh.ch/AgileAutonomy.html} and at~\cite{software_dataset}.

\section*{Supplementary materials}
\makebox[1.8cm][l]{Section S1.}  Sensitivity to noise in estimation and control. \\ 
\makebox[1.8cm][l]{Section S2.} The impact of obstacle density on performance.\\
\makebox[1.8cm][l]{Section S3.} The platform's hardware.\\
\makebox[1.8cm][l]{Section S4.} Computational complexity.\\
\makebox[1.8cm][l]{Section S5.} Theoretical maximum speed while avoiding a pole.\\
\makebox[1.8cm][l]{Section S6.} Metropolis-Hasting sampling.\\
\makebox[1.8cm][l]{Figure S1.} Sensitivity to noise in estimation and control.\\
\makebox[1.8cm][l]{Figure S2.} Ablation of tree density.\\
\makebox[1.8cm][l]{Figure S3.} Illustration of the experimental platform.\\
\makebox[1.8cm][l]{Figure S4.} Illustration of the simulated environments.\\
\makebox[1.8cm][l]{Figure S5.} Qualitative illustration of the related work.\\
\makebox[1.8cm][l]{Figure S6.} Motivation of global planning for label generation.\\
\href{}{\makebox[1.8cm][l]{\href{https://youtu.be/0doc6aMcRtU}{Movie S1}.} Deployment at different speeds.} \\
\href{}{\makebox[1.8cm][l]{\href{https://youtu.be/ALs60ij8JA8}{Movie S2}.} Deployment in simulation.} \\

\clearpage
\newpage
\appendix

\setcounter{table}{0}
\makeatletter 
\renewcommand{\thetable}{S\@arabic\c@table}
\makeatother

\setcounter{figure}{0}
\makeatletter 
\renewcommand{\thefigure}{S\@arabic\c@figure}
\makeatother

\setcounter{algorithm}{0}
\makeatletter 
\renewcommand{\thealgorithm}{S\@arabic\c@algorithm}
\makeatother

\setcounter{section}{0}
\makeatletter 
\renewcommand{\thesection}{S\@arabic\c@section}
\makeatother

\section*{Supplementary materials}

\section{Sensitivity to Estimation and Control Noise}\label{sec:noise_sensitivity}

We compare the sensitivity of our proposed approach to noise in state estimation as well as the applied control action against the \textit{Reactive} and \textit{FastPlanner} baselines. To this end, we test in the high-density simulated forest as explained in section~\ref{sec:sota_comp} and perturb the simulated state as well as the applied control command with additive disturbances.
Both types of disturbances are approximated as Gaussian noise with mean and variance identified from fast flights in an instrumented motion capture volume. 
This instrumented tracking volume allows us accurately identify noise and drift in state estimation and the applied control action.
Table~\ref{tab:noise_params} shows the identified noise parameters for the estimated state. 
The parameters are identified component-wise. 
Note that attitude is denoted in terms of the residual quaternion that rotates the estimated attitude onto the ground-truth attitude.
The noise in control action is represented as a multiplicative factor $m$ on the commanded collective thrust, which mimics a drop in motors efficiency because of aerodynamics effects or shortages in battery power.
This factor $m$ is selected from a uniform distribution $m \in \mathcal{U}(0.9,1)$ for each rollout.

As laid out in section~\ref{sec:sota_comp}, we test the performance for a set of increasing speeds, while the tree density is kept constant at the highest density of $\delta_3=\frac{1}{25}$ tree$\SI{}{\per\meter\squared}$. 
For each average forward speed, we perform $10$ rollouts and compute the respective success rate.
We repeat the set of experiments five times with different random seeds and report the mean and standard deviation overs sets of rollouts.

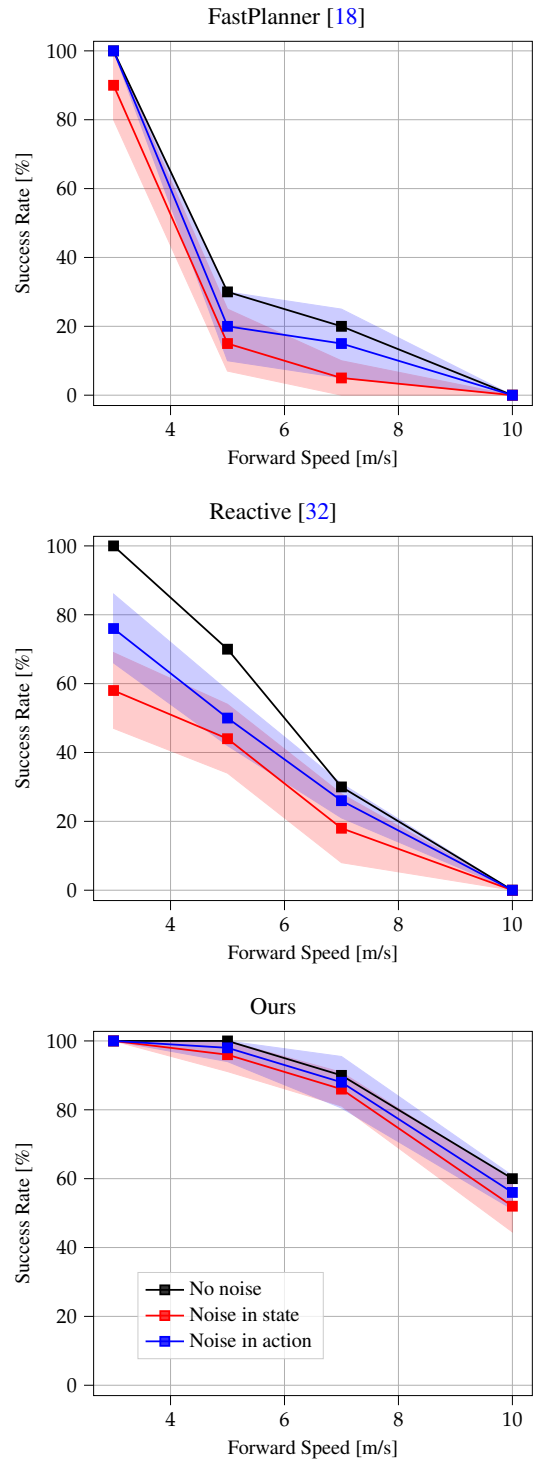
\begin{figure}[!h]
\centering
\setlength{\tabcolsep}{0pt}
\begin{tabular}{c}
\hspace{2ex} FastPlanner~\cite{zhou2019robust} \\
\resizebox{0.8\linewidth}{!}{\begin{tikzpicture}

\begin{axis}[
legend cell align={left},
legend style={at={(0.1,0.1)},anchor=south west,fill opacity=0.8, draw opacity=1, text opacity=1, draw=white!80!black},
tick align=outside,
tick pos=left,
x grid style={white!69.0196078431373!black},
xlabel={Forward Speed [m/s]},
xmajorgrids,
xmin=2.65, xmax=10.35,
xtick style={color=black},
y grid style={white!69.0196078431373!black},
ylabel={Success Rate [\%]},
ymajorgrids,
ymin=-3, ymax=102.774165738677,
ytick style={color=black}
]

\path [draw=red, fill=red, opacity=0.2]
(axis cs:3,100)
--(axis cs:3,100)
--(axis cs:5,25)
--(axis cs:7,10)
--(axis cs:10,0)
--(axis cs:10,0)
--(axis cs:10,0)
--(axis cs:7,0)
--(axis cs:5,7)
--(axis cs:3,80)
--cycle;

\path [draw=blue, fill=blue, opacity=0.2]
(axis cs:3,100)
--(axis cs:3,100)
--(axis cs:5,30)
--(axis cs:7,25)
--(axis cs:10,0)
--(axis cs:10,0)
--(axis cs:10,0)
--(axis cs:7,5)
--(axis cs:5,10)
--(axis cs:3,100)
--cycle;

\addplot [thick, black, mark=square*, mark size=2, mark options={solid}]
table {%
3 100
5 30
7 20
10 0
};
\addplot [thick, red, mark=square*, mark size=2, mark options={solid}]
table {%
3 90
5 15
7 5
10 0
};
\addplot [thick, blue, mark=square*, mark size=2, mark options={solid}]
table {%
3 100
5 20
7 15
10 0
};
\end{axis}

\end{tikzpicture}} \\
Reactive~\cite{florence2020integrated} \\
\resizebox{0.8\linewidth}{!}{\begin{tikzpicture}

\begin{axis}[
legend cell align={left},
legend style={at={(0.1,0.1)},anchor=south west,fill opacity=0.8, draw opacity=1, text opacity=1, draw=white!80!black},
tick align=outside,
tick pos=left,
x grid style={white!69.0196078431373!black},
xlabel={Forward Speed [m/s]},
xmajorgrids,
xmin=2.65, xmax=10.35,
xtick style={color=black},
y grid style={white!69.0196078431373!black},
ylabel={Success Rate [\%]},
ymajorgrids,
ymin=-3, ymax=102.774165738677,
ytick style={color=black}
]

\path [draw=red, fill=red, opacity=0.2]
(axis cs:3,69)
--(axis cs:3,69)
--(axis cs:5,54)
--(axis cs:7,28)
--(axis cs:10,0)
--(axis cs:10,0)
--(axis cs:10,0)
--(axis cs:7,8)
--(axis cs:5,34)
--(axis cs:3,47)
--cycle;

\path [draw=blue, fill=blue, opacity=0.2]
(axis cs:3,86)
--(axis cs:3,86)
--(axis cs:5,58)
--(axis cs:7,31)
--(axis cs:10,0)
--(axis cs:10,0)
--(axis cs:10,0)
--(axis cs:7,21)
--(axis cs:5,42)
--(axis cs:3,66)
--cycle;

\addplot [thick, black, mark=square*, mark size=2, mark options={solid}]
table {%
3 100
5 70
7 30
10 0
};
\addplot [thick, red, mark=square*, mark size=2, mark options={solid}]
table {%
3 58 
5 44 
7 18 
10 0 
};
\addplot [thick, blue, mark=square*, mark size=2, mark options={solid}]
table {%
3 76 
5 50 
7 26 
10 0 
};
\end{axis}

\end{tikzpicture}} \\
Ours \\
\resizebox{0.8\linewidth}{!}{\begin{tikzpicture}

\begin{axis}[
legend cell align={left},
legend style={at={(0.1,0.1)},anchor=south west,fill opacity=0.8, draw opacity=1, text opacity=1, draw=white!80!black},
tick align=outside,
tick pos=left,
x grid style={white!69.0196078431373!black},
xlabel={Forward Speed [m/s]},
xmajorgrids,
xmin=2.65, xmax=10.35,
xtick style={color=black},
y grid style={white!69.0196078431373!black},
ylabel={Success Rate [\%]},
ymajorgrids,
ymin=-3, ymax=102.774165738677,
ytick style={color=black}
]

\path [draw=red, fill=red, opacity=0.2]
(axis cs:3,100)
--(axis cs:3,100)
--(axis cs:5,91.1010205144336)
--(axis cs:7,81.1010205144336)
--(axis cs:10,44.5166852264521)
--(axis cs:10,59.4833147735479)
--(axis cs:10,59.4833147735479)
--(axis cs:7,90.8989794855664)
--(axis cs:5,100)
--(axis cs:3,100)
--cycle;

\path [draw=blue, fill=blue, opacity=0.2]
(axis cs:3,100)
--(axis cs:3,100)
--(axis cs:5,94)
--(axis cs:7,80.5166852264521)
--(axis cs:10,51.1010205144336)
--(axis cs:10,60.8989794855664)
--(axis cs:10,60.8989794855664)
--(axis cs:7,95.4833147735479)
--(axis cs:5,100)
--(axis cs:3,100)
--cycle;

\addplot [thick, black, mark=square*, mark size=2, mark options={solid}]
table {%
3 100
5 100
7 90
10 60
};
\addlegendentry{No noise}
\addplot [thick, red, mark=square*, mark size=2, mark options={solid}]
table {%
3 100
5 96
7 86
10 52
};
\addlegendentry{Noise in state}
\addplot [thick, blue, mark=square*, mark size=2, mark options={solid}]
table {%
3 100
5 98
7 88
10 56
};
\addlegendentry{Noise in action}
\end{axis}

\end{tikzpicture}} \\

  \end{tabular}
\caption{\textbf{Sensitivity analysis} of noise in state estimation and control.}\label{fig:noise_ablation}
\end{figure}

Figure~\ref{fig:noise_ablation} shows the results of this experiment, with the line indicating the mean success rate and the shaded area marking one standard deviation. 
At a low speed of $\SI{3}{\meter\per\second}$, the performance of our approach is not affected by neither the noise in estimation nor control. 
In comparison, the \textit{Reactive} baseline shows significant drop in performance when exposed to noise in action and even more in case of noise in state estimation. 
The \textit{FastPlanner} is less sensitive to noise and only exhibits a small drop in performance at $\SI{3}{\meter\per\second}$.
For higher speeds, the performance of our approach with noise is between 5\% and 10\% worse than in the noise-free case, with noise in estimation resulting in a larger drop in success rate. 
The same pattern can be observed for the \textit{FastPlanner} and the \textit{Reactive} baselines, which both show a larger performance decrease when exposed to noise in estimation. 
In summary, neither noise in estimation nor noise in control did significantly affect performance of our approach, while the performance of the \textit{Reactive} baseline is strongly reduced. The \textit{FastPlanner} baseline shows larger robustness against noise, but performs inferior in general.

\section{Performance Analysis in Function of \\ Obstacle Density}
\label{sec:performance_vs_density}
\begin{figure*}
    \centering
\begin{tabular}{c|c c c | c c c | c c c c }
\toprule
  &$v_x$ & $v_y$ & $v_z$ & $\omega_x$ & $\omega_y$ & $\omega_z$ & $q_w$ & $q_x$ & $q_y$ & $q_z$ \\
  \midrule
  $\mu$ & $0.009$ & $-0.198$ & $-0.570$ & $-0.009$ & $0.012$ & $-0.004$ & $0.997$ & $0.002$ & $0.022$ & $0.003$ \\
  $\sigma$ & $0.496$ & $0.210$ & $1.243$ & $0.302$ & $0.587$ & $0.031$ & $0.000$ &  $0.003$ & $0.001$ & $0.001$ \\
  \bottomrule
\end{tabular}
\captionof{table}{\textbf{State estimation noise} for velocity, angular velocity and attitude (quaternion). The noise parameters are obtained from real-world flights in a vicon-based system.\label{tab:noise_params}}
\centering
\def\colwidth{0.33\textwidth}
\def\doublecolwidth{0.66\textwidth}
\newcolumntype{M}[1]{>{\centering\arraybackslash}m{#1}}
\addtolength{\tabcolsep}{-4pt}
\begin{tabular}{M{\colwidth} m{0.7em} M{\doublecolwidth}}
\vspace{-2ex}
\resizebox{0.9\linewidth}{!}{%
\begin{tikzpicture}

\begin{axis}[
legend style={at={(0,1.1)},anchor=south west, /tikz/every even column/.append style={column sep=0.093cm}},
legend columns=2,
legend cell align={left},
tick align=outside,
tick pos=left,
x grid style={white!69.01960784313725!black},
xmajorgrids,
xmin=2.55, xmax=12.45,
xtick style={color=black},
y grid style={white!69.01960784313725!black},
ylabel={Success Rate [\%]},
ymajorgrids,
ymin=-5, ymax=105,
ytick style={color=black}
]
\addplot [thick, blind, mark=square*, mark size=2, mark options={solid}]
table {%
3 30
5 30
7 30
10 30
12 30
};
\addlegendentry{Blind}
\addplot [thick, fastplanner, mark=triangle*, mark size=3, mark options={solid}]
table {%
3 100
5 30
7 30
10 0
12 0
};
\addlegendentry{FastPlanner~\cite{zhou2019robust}}
\addplot [thick, reactive, mark=triangle*, mark size=3, mark options={solid,rotate=180}]
table {%
3 100
5 90
7 30
10 0
12 0
};
\addlegendentry{Reactive~\cite{florence2020integrated}}
\addplot [thick, ours, mark=*, mark size=3, mark options={solid}]
table {%
3 100
5 100
7 100
10 90
12 50
};
\addlegendentry{Ours}
\end{axis}

\end{tikzpicture}%
}& 
\vspace{3ex}
\begin{turn}{90}
$\delta_1=1/49$
\end{turn} &
\vspace{3ex}
\includegraphics[trim=0 0 0 150, clip,width=0.95\linewidth]{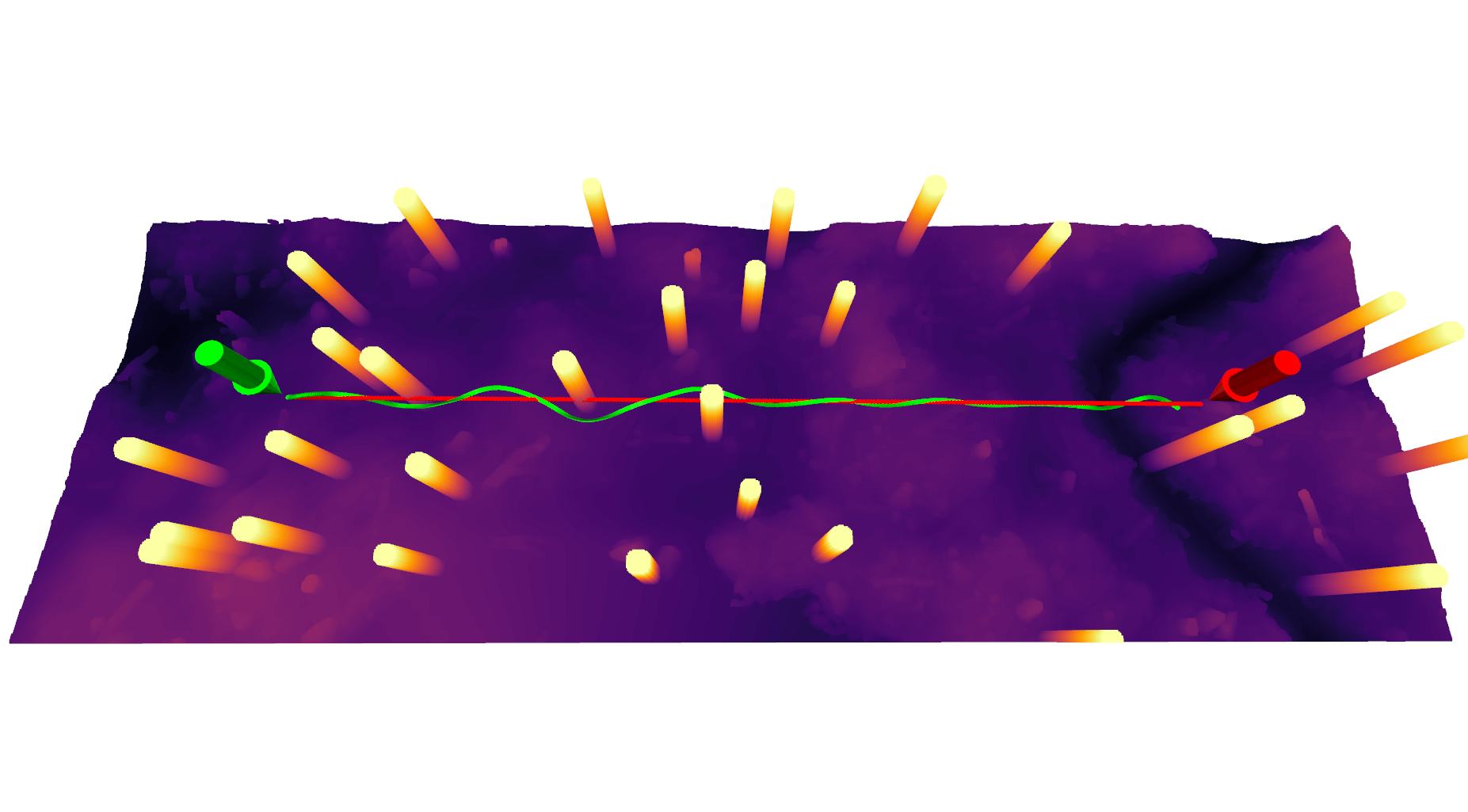} \tabularnewline
\vspace{-7ex}
\resizebox{0.9\linewidth}{!}{%
\begin{tikzpicture}

\begin{axis}[
tick align=outside,
tick pos=left,
x grid style={white!69.01960784313725!black},
xmajorgrids,
xmin=2.55, xmax=12.45,
xtick style={color=black},
y grid style={white!69.01960784313725!black},
ylabel={Success Rate [\%]},
ymajorgrids,
ymin=-5, ymax=105,
ytick style={color=black}
]
\addplot [thick, blind, mark=square*, mark size=2, mark options={solid}]
table {%
3 20
5 20
7 20
10 20
12 20
};
\addplot [thick, fastplanner, mark=triangle*, mark size=3, mark options={solid}]
table {%
3 100
5 20
7 20
10 0
12 0
};
\addplot [thick, reactive, mark=triangle*, mark size=3, mark options={solid,rotate=180}]
table {%
3 100
5 80
7 10
10 0
12 0
};
\addplot [thick, ours, mark=*, mark size=3, mark options={solid}]
table {%
3 100
5 100
7 100
10 70
12 40
};
\end{axis}

\end{tikzpicture}%
}&
\vspace{-7ex}
\begin{turn}{90}
$\delta_2=1/36$
\end{turn} &
\vspace{-7ex}
\includegraphics[trim=0 0 0 150, clip,width=0.95\linewidth]{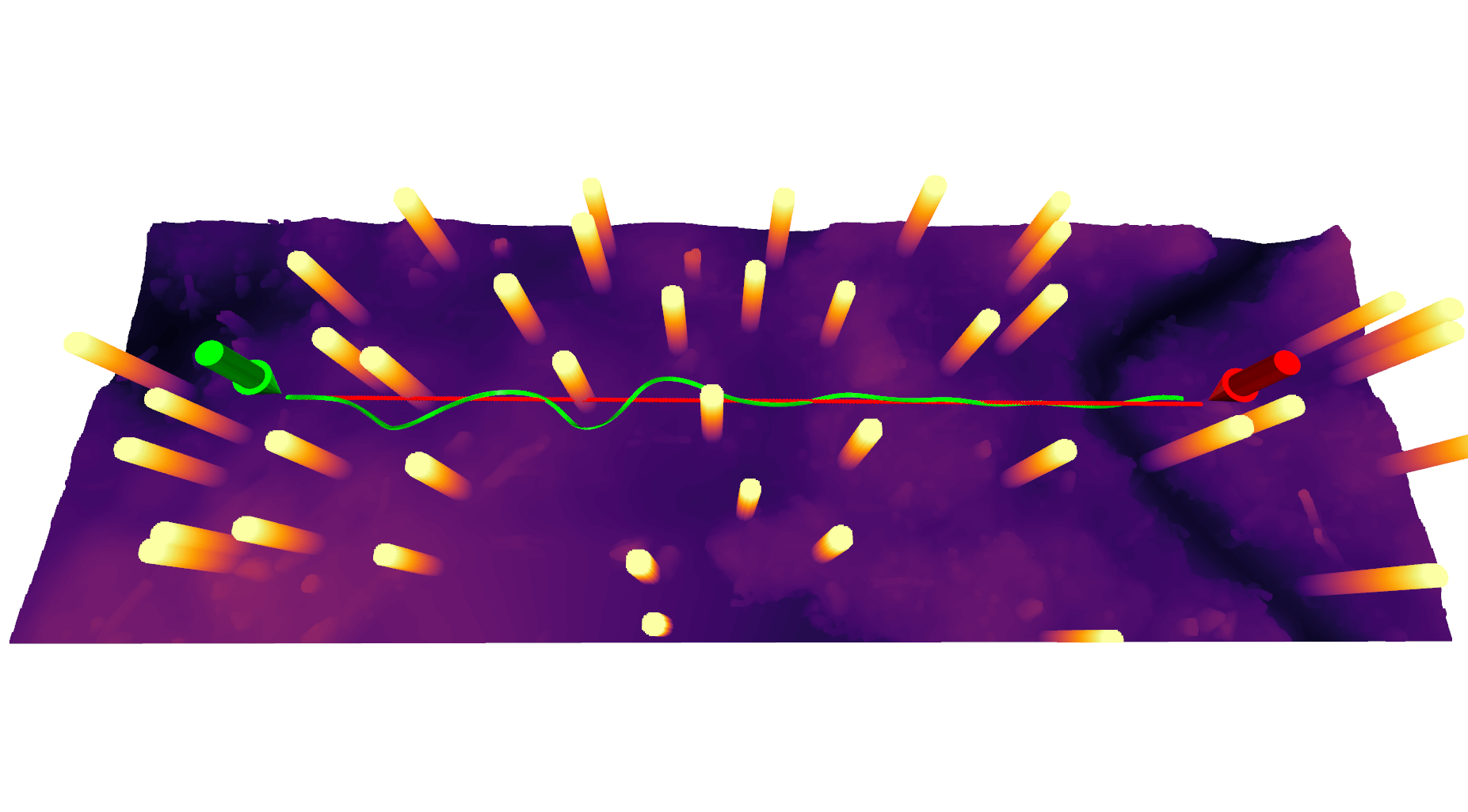} \tabularnewline
\vspace{-7ex}
\resizebox{0.9\linewidth}{!}{%
\begin{tikzpicture}

\begin{axis}[
legend style={at={(0,1.1)},anchor=south west, /tikz/every even column/.append style={column sep=0.093cm}},
legend columns=2,
legend cell align={left},
tick align=outside,
tick pos=left,
x grid style={white!69.01960784313725!black},
xmajorgrids,
xmin=2.55, xmax=10.45,
xtick style={color=black},
y grid style={white!69.01960784313725!black},
ylabel={Success Rate [\%]},
ymajorgrids,
ymin=-5, ymax=105,
ytick style={color=black}
]
\addplot [thick, blind, mark=square*, mark size=2, mark options={solid}]
table {%
3 20
5 20
7 20
10 20
};
\addlegendentry{Blind}
\addplot [thick, fastplanner, mark=triangle*, mark size=3, mark options={solid}]
table {%
3 100
5 20
7 20
10 0
};
\addlegendentry{FastPlanner}
\addplot [thick, reactive, mark=triangle*, mark size=3, mark options={solid,rotate=180}]
table {%
3 100
5 70
7 30
10 0
};
\addlegendentry{Reactive}
\addplot [thick, ours, mark=*, mark size=3, mark options={solid}]
table {%
3 100
5 100
7 90
10 60
};
\addlegendentry{Ours}
\end{axis}

\end{tikzpicture}%
}&
\vspace{-9ex}
\begin{turn}{90}
$\delta_3=1/25$
\end{turn} &
\vspace{-7ex}
\includegraphics[trim=0 0 0 150, clip,width=0.95\linewidth]{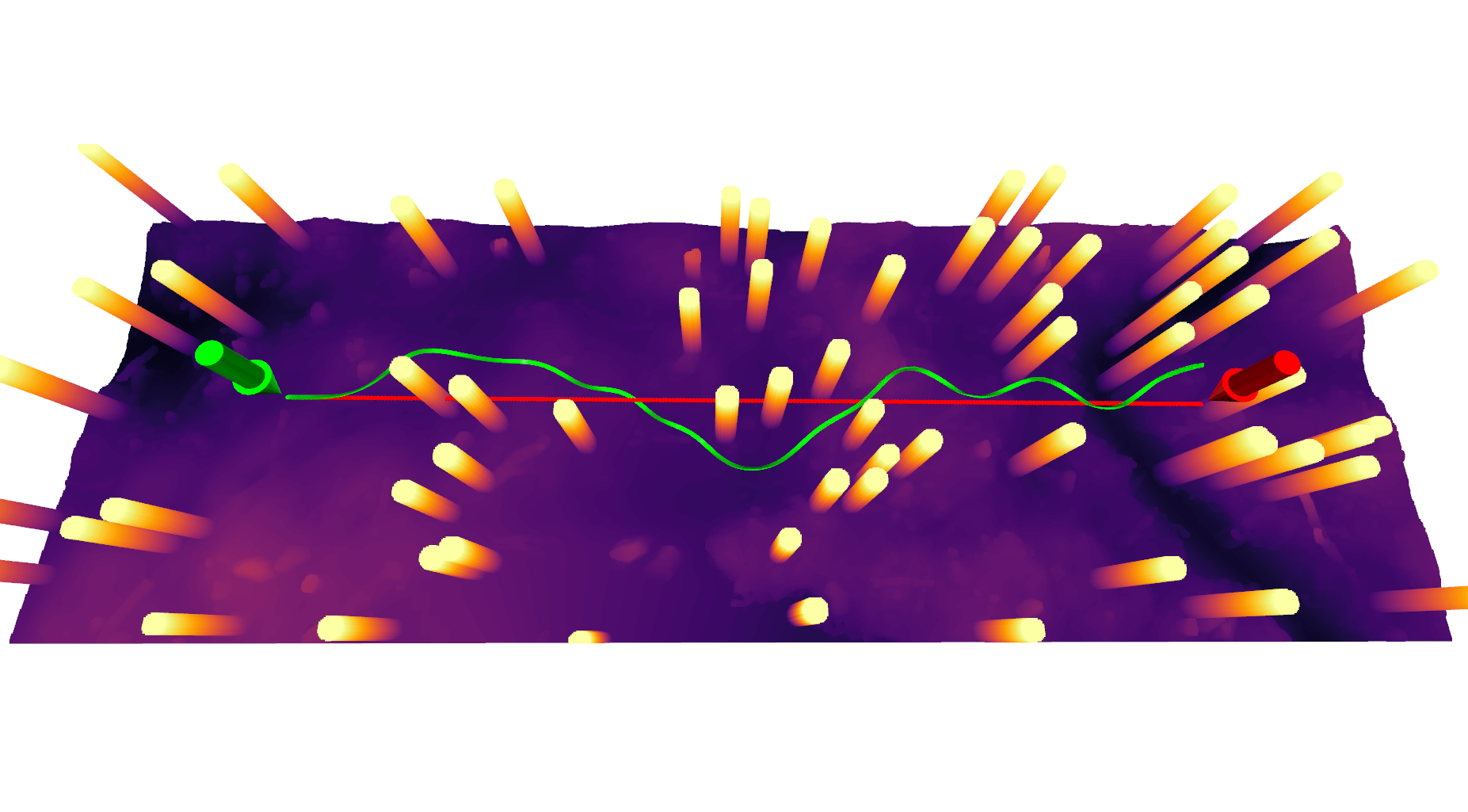} \tabularnewline
\end{tabular}
\addtolength{\tabcolsep}{4pt}
\captionof{figure}{\textbf{Experiments in a simulated forest}. Experiments are ordered for increasing difficulty, which is controlled by the tree density $\delta$. The left column reports success rates at various speeds. The right column shows one of the random realizations of the environment together with paths taken by different policies from start (green arrow) to end (red arrow). The paths illustrate the blind policy (red) and the path taken by our approach (green). Our approach consistently outperforms the baselines in all environments and and at all speeds. \label{fig:comparison_density}}
 \vspace{-4ex}
\end{figure*}

\begin{figure*}[t]
    \centering
    \def\cwidth{0.49\textwidth}
    \newcolumntype{M}[1]{>{\centering\arraybackslash}m{#1}}
    \addtolength{\tabcolsep}{-4pt}
    \begin{tabular}{M{\cwidth} M{\cwidth}}
    \includegraphics[width=\linewidth]{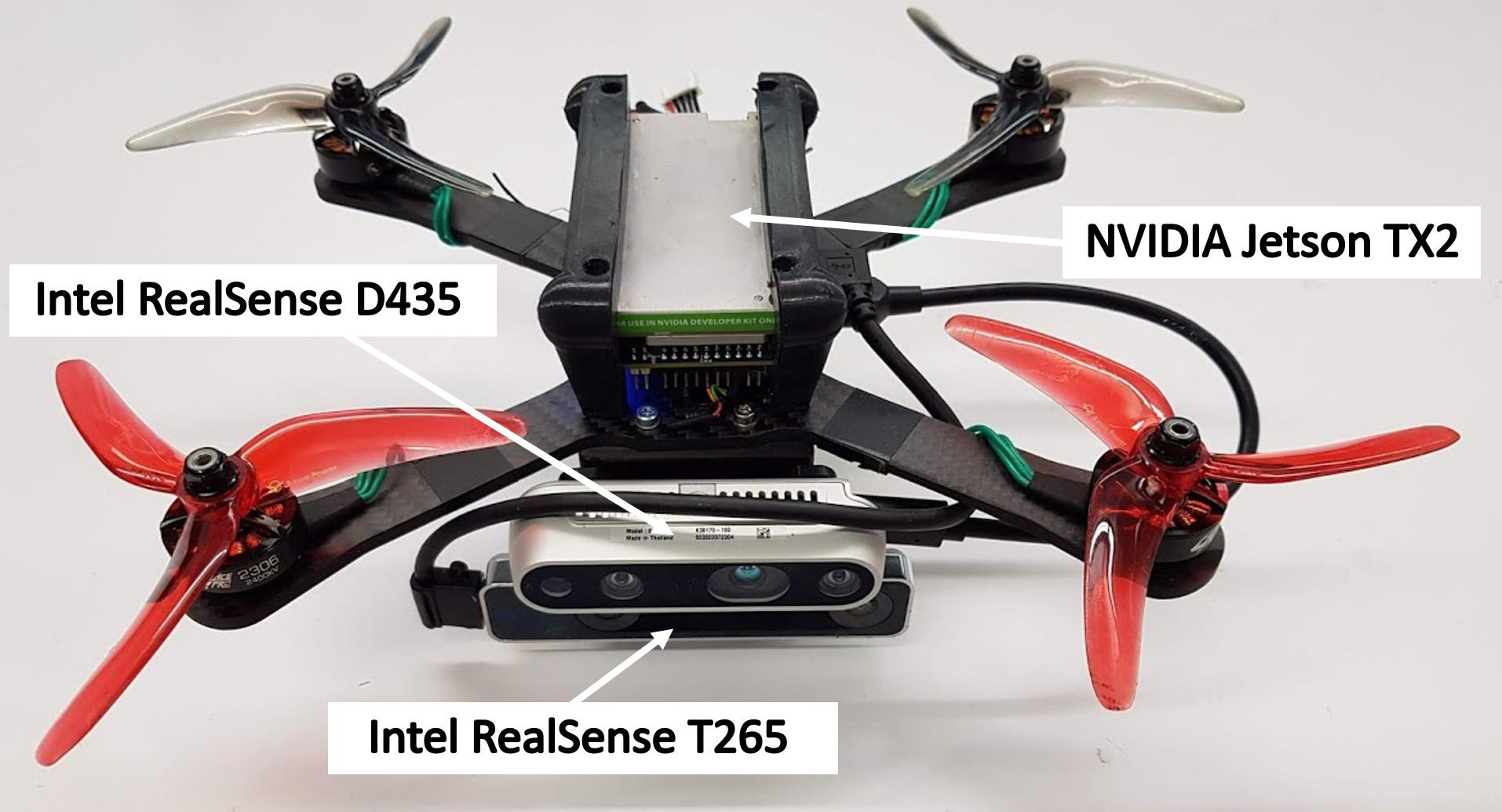} &
    \includegraphics[width=\linewidth]{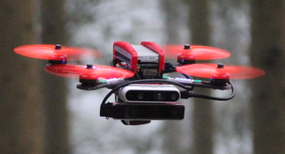} \\
\end{tabular}
    \caption{\textbf{ Illustration of our experimental platform}. The main computational unit is an NVIDIA Jetson TX2, whose GPU is used for neural network inference and CPU for the control stack. Sensing is performed by an Intel Realsense T265 for state estimation and an Intel Realsense D435 for depth estimation.~\label{fig:drone_illustration}}
\end{figure*}

We perform a set of experiments in a simulated forest to evaluate the performance of our approach with respect to obstacle density.
Similarly to previous experiments, we use as baselines the mapping and planning method of Zhou et al.~\cite{zhou2019robust} (\emph{FastPlanner}) and the reactive planner of Florence et al.~\cite{florence2020integrated} (\emph{Reactive}).
We compare the different approaches according to their success rate, which measures how often the drone reaches the goal location within a radius of 5 meters without crashing.

We build a simulated forest~\cite{karaman2012high} in a rectangular region $R(l,w)$ of width $w$ and length $l$, and fill it with trees that have a diameter of about \SI{0.6}{\meter}. Trees are randomly placed according to a homogeneous Poisson point process $P$ with intensity $\delta$~tree$\SI{}{\per\meter\squared}$~\cite{karaman2012high}. 
Note that Tomppo et al.~\cite{tomppo1986models} found that around 30\% of the forests in Finland could be considered as a realization of a spatial Poisson point process.
We control the task difficulty by changing the tree density $\delta$.
We set $w=\SI{30}{\meter}$ and $l=\SI{60}{\meter}$ and start the drone at position $s=(-\frac{l}{2}, -\frac{w}{2})$ (the origin of the coordinate system is at the center of $R(l, w)$).
We provide the drone with a straight reference trajectory of \SI{40}{\meter} length.
We test on three different tree densities with increasing difficulty: $\delta_1=\frac{1}{49}$ (\emph{low}), $\delta_2=\frac{1}{36}$ (\emph{medium}), and $\delta_3=\frac{1}{25}$ (\emph{high})~tree$\SI{}{\per\meter\squared}$ .
We vary the average forward speed of the drone between \SI{3}{\meter\per\second} and \SI{12}{\meter\per\second}.
We repeat the experiments with $10$ different random realizations of the forest for each difficulty, using the same random seed for all methods.

The first three rows in Figure~\ref{fig:comparison_density} show the results of this experiment, together with one example environment for each difficulty.
At a low speed of \SI{3}{\meter\per\second}, all methods successfully complete every run even for high difficulties.
As speed increases, the success rates of the baselines quickly degrade. At \SI{10}{\meter\per\second}, no baseline completes even a single run successfully, irrespective of task difficulty.
In contrast, our approach is significantly more robust at higher speeds. It achieves 100\% success rate 
up to \SI{5}{\meter\per\second}. For speeds of \SI{10}{\meter\per\second}, our approach has a success rate of 90\% in the low difficulty task and 60\% in the high difficulty task. Moreover, 
we show that our approach can go as fast as \SI{12}{\meter\per\second} with a success rate of up to 50\%.
Our method achieves this performance by decreasing the latency between sensors and actions and increasing the robustness to sensor noise, developed using regularities in the data.
We show additional controlled studies on latency and sensor noise in section~\ref{sec:computational_cost} and section~\ref{sec:noise_robustness}.

\section{Experimental Platform} \label{sec:experimental_platform}

To validate our approach with real-world experiments, we designed a lightweight, but powerful, quadrotor platform.
The main frame is an Armattan Chameleon $\SI{6}{\inch}$, equipped with Hobbywing XRotor 2306 motors and $\SI{5}{\inch}$, three-bladed propellers.
The platform has a total weight of $890$ grams and can produce a maximum thrust of about \SI{40}{\newton}, which results in a thrust-to-weight ratio of $4.4$.
The weight and power of this platform is comparable to the ones used by professional pilots in drone racing competitions.

The platform's main computational unit is an NVIDIA Jetson TX2 accompanied by a ConnectTech Quasar carrier board.
The GPU of the Jetson TX2 is used to run neural network inference and its CPU for the rest of our control framework.
The output of this framework is a low-level control command including a collective thrust and angular rates to be achieved for flying.
The desired commands are sent to a commercial flight controller running BetaFlight, which produces single-rotor commands that are fed to the 4-in-1 motor controller.

Our quadrotor is equipped with two off-the-shelf sensing units: an Intel RealSense T265 and an Intel RealSense D435i.
Both have a stereo camera setup and an integrated IMU.
The RealSense T265 runs a visual-inertial odometry pipeline to output the state estimation of the platform at $\SI{200}{\hertz}$, which we directly use without any additional processing.
Our second sensing unit, the RealSense D435i, outputs a hardware-accelerated depth estimation pipeline on its stereo setup.
The latter pipeline provides to our framework a dense depth in VGA resolution ($\SI[product-units = single]{640x480}{px}$) at $\SI{30}{\hertz}$, which we use without further processing.
The horizontal field of view of this observation is about $\SI{90}{\degree}$, which appeared to be sufficient for the task of obstacle avoidance.
For experiments at speeds of \SI{7}{\meter\per\second} and above, we tilt the depth sensor by \SI{30}{\degree} to assure that the camera would look forward during flight.
This is indeed a typical camera setup for high-speed flight in drone racing competitions.
To support the transfer from simulation to reality, we build a simulated stereo setup with the same characteristics of the RealSense D435, on which we run a \href{https://github.com/dhernandez0/sgm}{GPU implementation of SGM}~\cite{sgm_gpu_iccs2016} to estimate dense depth in simulation.

Our software stack is developed in both C++ and Python, and will be made publicly available upon acceptance.
Specifically, we implement the trajectory prediction framework with Tensorflow in a Python node and the rest of our trajectory projection and tracking software in separate C++ nodes.
The communication between different nodes is implemented with ROS.
Specifically, the Tensorflow node predicts trajectories at \SI{24.7}{\hertz} on the physical platform, and the MPC generates commands in the form of collective thrust and body rates at $\SI{100}{\hertz}$ to track those trajectories.
The low-level controller, responsible for tracking desired body rates and collective thrust predicted by the MPC, runs at \SI{2}{\kilo\hertz}.
The platform only receives a start and stop command from the base computer, and is therefore completely autonomous during flight.

\section{Computational Complexity}~\label{sec:computational_cost_supp}

The baseline with the largest latency is FastPlanner.
This baseline has three components: sensing, mapping and planning. Sensing includes transforming a depth image to a pointcloud after filtering. Mapping includes ray casting and Euclidean Signed Distance Field (ESDF) computation.
Finally, planning includes finding the best trajectory to reach the goal while avoiding obstacles.
The total time to perform all these operations is \SI{65.2}{\milli\second}. 
However, it is important to note that, while the depth filtering and the probabilistic ray casting process are necessary to remove sensing errors, those operations make the inclusions of obstacles in the local map slower. 
Practically, 2-3 observations are required to add an obstacle to the local map, therefore largely increasing the overall latency of the system.

Removing the mapping stage altogether, the \textit{Reactive} baseline experiences significant gains in computation time.
For this baseline, the sensing latency is the time between receiving a depth observation and generating a point cloud after filtering and outlier rejection.
The planning latency consists of building a KD-Tree from the filtered point cloud, and selecting the best trajectory out of the available motion primitives according to a cost based on collision probability and proximity to the goal.
This baseline is about three times faster than \textit{FastPlanner}, with a total latency of \SI{19.1}{\milli\second}.
However, the reduced latency comes at the cost of a lower trajectory-representation power, since the planner can only select a primitive from a pre-defined motion library.
In addition, given the lack of temporal filtering, the reactive baseline is very sensitive to sensing errors, which can drastically affect performance at high speeds.

Our approach has significantly lower latency than both baselines: when network inference is performed on the GPU, our approach is 25.3 times faster than \textit{FastPlanner} and 7.4 times faster than the \textit{Reactive} baseline.
When GPU inference is disabled, the network's latency increases by only \SI{8}{\milli\second}, and our approach is still significantly faster than both baselines.
For our approach, we break down computation into three operations: (i) sensing, which includes the time for recording an image and convert it to an input tensor, (ii) neural network inference, and (iii) projection, which is the time to project the prediction of the neural network into the space of dynamically feasible trajectories for the quadrotor.
Moving from the desktop computer to the onboard embedded computing device, the network's forward pass requires \SI{38.9}{\milli\second}.
Onboard, the total time to pass from sensor to action is then \SI{41.6}{\milli\second}, which is sufficient to update actions at up to \SI{24.3}{\hertz}.

\section{Rotational Dynamics}~\label{sec:rotation_effects}

Extending~\cite{Falanga19ral_howfast} to the quadrotor platform, we approximate the maximum speed $v_{\text{max}}$ that still allows successful avoidance of a vertical cylindrical obstacle.
The avoidance maneuver is described as a sequence of two motion primitives consisting of pure rolling and pure acceleration. 
Both primitives are executed with maximum motor inputs. 
Concretely, we treat the time required to reorient the quadrotor as additional latency in the system, which can be computed by
\begin{equation}\label{eq:rot_latency}
    t_{\text{rot}} = \sqrt{\frac{2\phi J}{T_\text{max}}} ,
\end{equation}
with $J$ being the moment of inertia, $T_\text{max}$ the maximum torque the quadrotor can produce, and $\phi$ the desired roll angle. 
As the roll angle becomes a decision variable itself in this setting, we identify the best roll angle by maximizing the speed that still allows successful avoidance.
Assuming that the quadrotor oriented at a roll angle $\phi$ accelerates with full thrust, its lateral position $p_\text{lat}$ can be described by
\begin{equation}
    p_\text{lat}(t) = \frac{1}{2} \sin{\phi} \cdot c_{\text{max}} \cdot t^2 \; ,
\end{equation}
where $c_\text{max}$ denotes the maximum mass-normalized thrust of the platform. 
Solving this equation for $t$ and setting $p_\text{lat} = r_{\text{obs}}$ allows to formulate a maximum linear speed that still allows successful avoidance when considering the sensing range $s$, the combined radius of the obstacle and the drone $r_{\text{obs}}$, the sensing latency $t_{\text{s}}$, the processing latency $t_{\text{p}}$, \ie the time to convert an observation into motor commands, and the latency introduced to reorient the platform $t_{\text{rot}}$:
\begin{equation}\label{eq:max_speed}
    v_{\text{max}} = \frac{s}{t_{\text{s}} + t_{\text{p}} + t_{\text{rot}} + \sqrt{\frac{2r_{\text{obs}}}{\sin{\phi} \cdot c_{\text{max}}}}} \; .
\end{equation}

Inserting \ref{eq:rot_latency} into \ref{eq:max_speed} we can identify $\phi$ that maximizes $v_\text{max}$.
In our study case, we avoid a pole with diameter $\SI{1.5}{\meter}$ and model the drone as a sphere with diameter $\SI{0.4}{\meter}$. Therefore, the combined radius of the obstacle and drone is $r_{\text{obs}} = \SI{0.95}{\meter}$.
In addition, we set $c_\text{max} = \SI{35.3}{\meter\per\second\squared}$, $J=\SI{0.007}{\kilogram\per\meter\squared}$.
Similarly to Falanga et al.~\cite{Falanga19ral_howfast}, we define the sensing latency as the worst-case time between the quadrotor being closer than the sensing range $s$ to the obstacle and an image containing the obstacle being rendered and provided to the navigation algorithms.
This time corresponds to the inverse of the frame rate.
Therefore, $t_{\text{s}}=\SI{66}{\milli\second}$, since images are rendered at \SI{15}{\hertz}.
Also similarly to Falanga et al.~\cite{Falanga19ral_howfast}, we select as sensing range $s$ for our stereo camera the range at which the depth uncertainty because of triangulation is below 20\% of
the actual depth.
Using this definition and the parameters of our simulated stereo camera ( VGA resolution, baseline of $\SI{0.1}{\meter}$, and focal length of \SI{4}{\milli\meter}), we compute a sensing range of $s=\SI{6}{\meter}$.
Table~\ref{tab:sensing_params} summarizes all the variables of the problem.
\begin{table}[]
    \centering
    \resizebox{\linewidth}{!}{
    \begin{tabular}{c c c c c c }
    \toprule
         $s$ & $T_{\text{max}}$ & $J$ & $c_{\text{max}}$ & $t_{\text{s}}$ & $r_{\text{obs}}$ \\
         \midrule
         $\SI{6}{\meter}$ & $\SI{1.02}{\newton\meter}$ &  $\SI{0.007}{\kilogram\per\meter\squared}$ & $\SI{35.3}{\meter\per\second\squared}$ & $\SI{66}{\milli\second}$ & $\SI{0.95}{\meter}$ \\
         \bottomrule
    \end{tabular}
    }
    \caption{Fixed parameters for the calculation of the theoretical maximum speed of our simulated sensor and quadrotor. Note that, differently from the physical quadrotor, the simulated drone is in the \emph{plus} configuration.}
    \label{tab:sensing_params}
\end{table}

Using these values, we derive the theoretical maximum speed for each method by considering their processing latency $t_{\text{p}}$ (c.f. section~\ref{sec:computational_cost}) and optimizing Equation~\ref{eq:max_speed} according to the free variable $\phi$.
Table~\ref{tab:max_speed} presents the results of the optimization.
\begin{table}[]
    \centering
    \begin{tabular}{c c c c c}
    \toprule
         & $t_{\text{p}}$ & $t_{\text{rot}}$ & $\phi$ & $v_{\text{max}}$ \\
         \midrule
         FastPlanner~\cite{zhou2019robust} &\SI{65.2}{\milli\second} & $\SI{125.2}{\milli\second}$  & $\SI{65.5}{\degree}$ & $\SI{12.0}{\meter\per\second}$   \\
         Reactive~\cite{zhou2019robust} &\SI{19.1}{\milli\second} & $\SI{125.2}{\milli\second}$  & $\SI{65.5}{\degree}$ & $\SI{13.2}{\meter\per\second}$    \\
         Ours & \SI{10.3}{\milli\second} & $\SI{125.2}{\milli\second}$  & $\SI{65.5}{\degree}$ & $\SI{13.5}{\meter\per\second}$ \\
         \bottomrule
    \end{tabular}
    \caption{We compute the theoretical maximum speed by optimizing Equation~\ref{eq:max_speed} according to the variable $\phi$ using grid-search. The processing latency $t_{\text{p}}$ represents the time to convert an image to a motor command (see section~\ref{sec:computational_cost}), and the rotation latency $t_{\text{rot}}$ is the time for the quadrotor to rotate at the angle $\phi$. Interestingly, the methods' processing latencies do not affect the rotation angle and latency.}
    \label{tab:max_speed}
\end{table}
Note that this approximation does not account for the fact that the quadrotor platform already performs some lateral acceleration during the rotation phase.

\section{Metropolis-Hastings Sampling} \label{sec:metropolis}
In statistics, the Metropolis-Hastings (M-H) algorithm~\cite{hastings70} is used to sample a distribution $P(w)$ which can't be directly accessed.
To generate the samples, the M-H algorithm requires a score function $d(w)$ proportional to $P(w)$.
Requiring $d(w) \propto P(w)$ waives the need to find the normalization factor $Z=\int_w d(w)$ such that $P(w) = \frac{1}{Z} d(w)$, which can't be easily calculated for high-dimensional spaces.
Metropolis-Hastings is a Markov Chain Monte Carlo sampling method~\cite{AndrieuFDJ03}.
Therefore, it generates samples by constructing a Markov chain that has the desired distribution as its equilibrium distribution.
In this Markov chain, the next sample $w_{t+1}$ comes from a distribution $t(w_{t+1} | w_t)$, referred to as transition model, which only depends on the current sample $w_t$.
The transition model $t(w_{t+1} | w_t)$ is generally a pre-defined parametric distribution, e.g. a Gaussian.
The next sample $w_{t+1}$ is then accepted and used for the next iteration, or it is rejected, discarded, and the current sample $w_t$ is re-used.
Specifically, the sample is accepted with probability equal to 
\begin{equation}
    \alpha = \min\left(1, \frac{d(w_{t+1})}{d(w_t)}\right) = \min\left(1, \frac{P(w_{t+1})}{P(w_t)}\right).
\end{equation}
Therefore, M-H always accepts a sample with a higher score than its predecessor.
However, the move to a sample with a smaller score will sometimes be rejected, and the higher the drop in score $\frac{1}{\alpha}$, the smaller the probability of acceptance.
Therefore, many samples come from the high-density regions of $P(w)$, while relatively few from the low-density regions.
Roberts et. al~\cite{roberts1994simple} have shown that under the mild condition that
\begin{equation}\label{eq:condition_mh}
    \alpha > 0 \quad \forall \quad w_t, w_{t+1} \in \mathcal{W}, 
\end{equation}
$\hat{P}(w)$ will asymptotically converge to the target distribution $P(w)$.
According to Eq.~\ref{eq:condition_mh}, the probability of accepting a sample with lower score than its predecessor is always different from zero, which implies that the method will not ultimately get stuck into a local extremum.
Intuitively, this is why the empirical sample distribution $\hat{P}(w)$ approximates the target distribution $P(w)$.
In contrast to other Monte Carlo statistical methods, e.g. importance sampling, the MH algorithm tend to suffer less from the curse of dimensionality and are therefore preferred for sampling in high-dimensional spaces~\cite{roberts1994simple}.

\begin{figure*}[!h]
\def\colwidth{0.315\linewidth}
\newcolumntype{M}[1]{>{\arraybackslash}m{#1}}
\addtolength{\tabcolsep}{-2pt}
\begin{tabular}{M{\colwidth} M{\colwidth} M{\colwidth}}
\textbf{A} Forest &  &  \tabularnewline
\includegraphics[width=\linewidth]{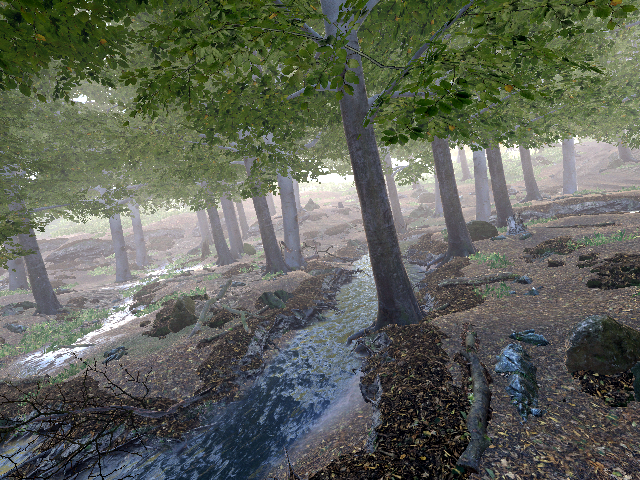} &
\includegraphics[width=\linewidth]{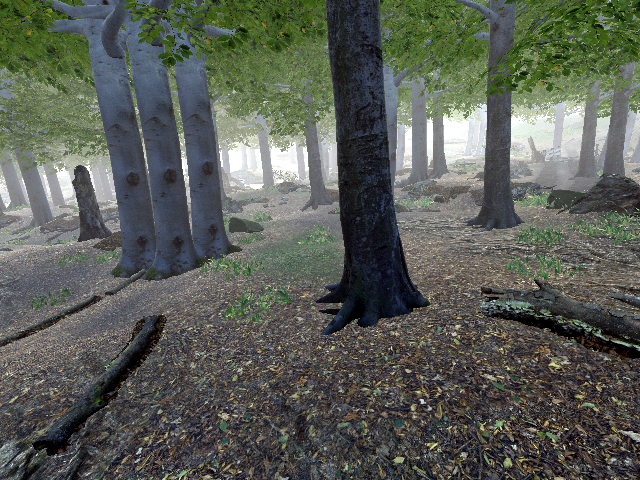} &
\includegraphics[width=\linewidth]{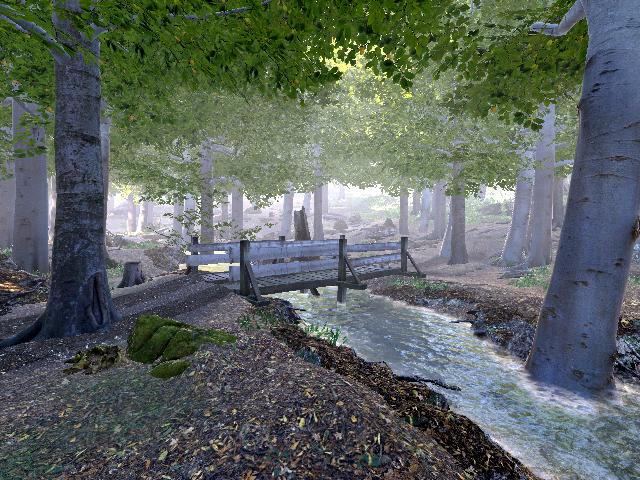} \tabularnewline
\textbf{B} Objects &  & \textbf{C} Narrow Gap  \tabularnewline
\includegraphics[width=\linewidth]{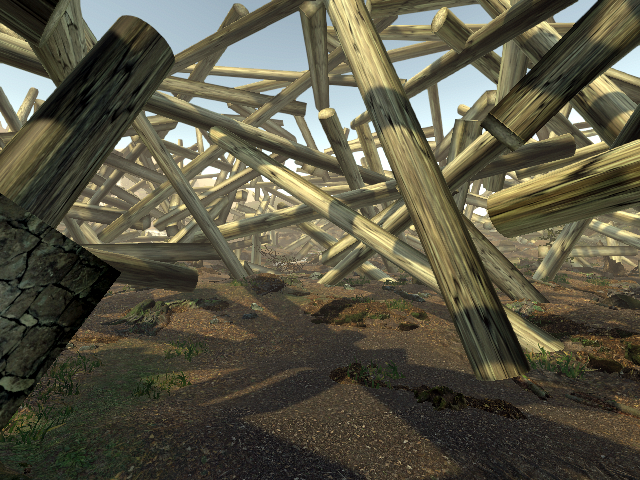} &
\includegraphics[width=\linewidth]{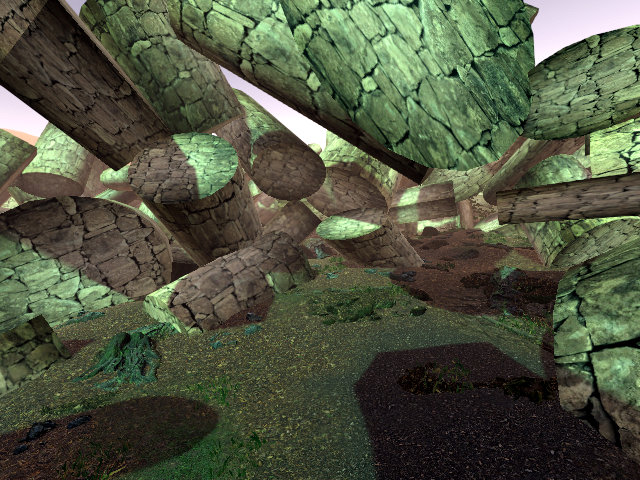} &
\includegraphics[width=\linewidth]{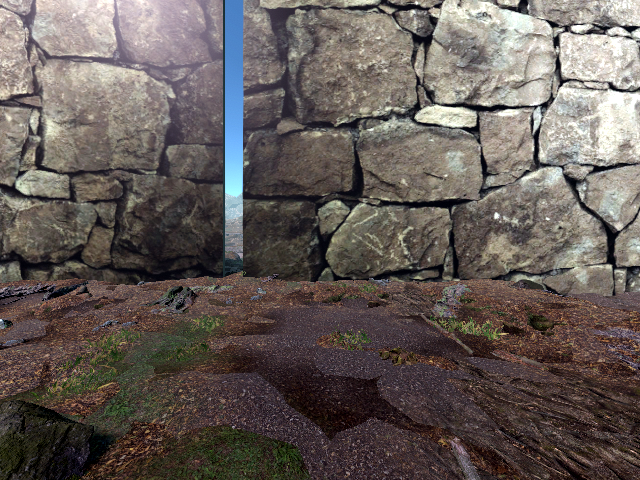} \tabularnewline
\textbf{D} Disaster &  &  \tabularnewline
\includegraphics[width=\linewidth]{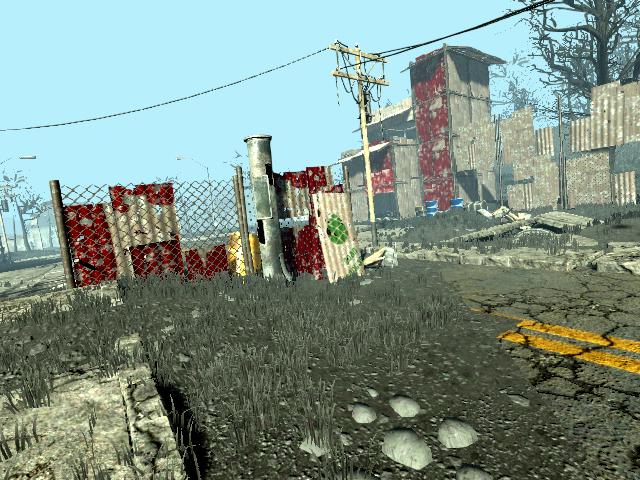} &
\includegraphics[width=\linewidth]{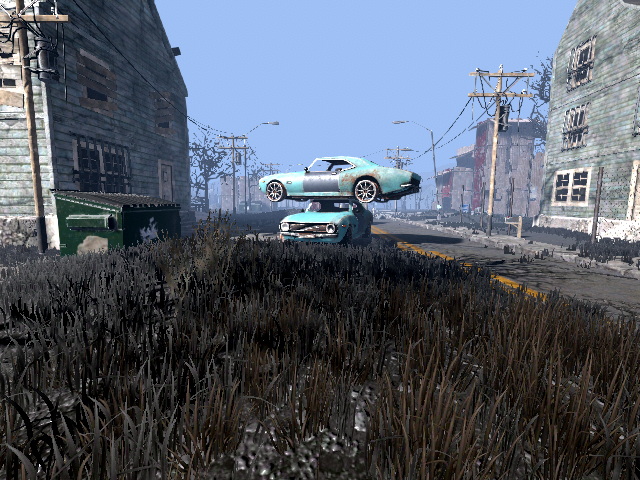} &
\includegraphics[width=\linewidth]{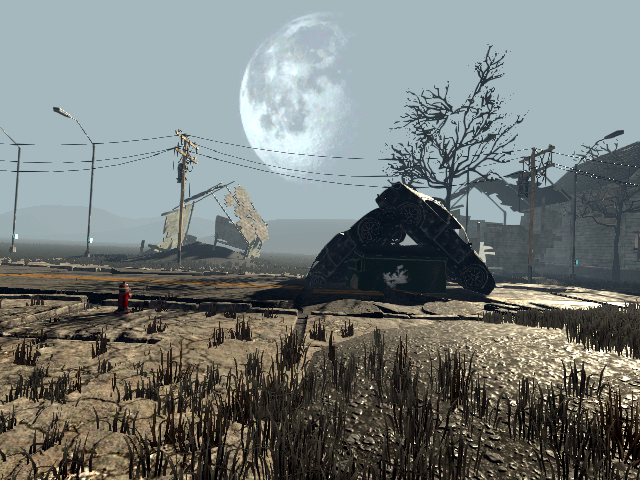} \tabularnewline
\textbf{E} City Street &  &  \tabularnewline
\includegraphics[width=\linewidth]{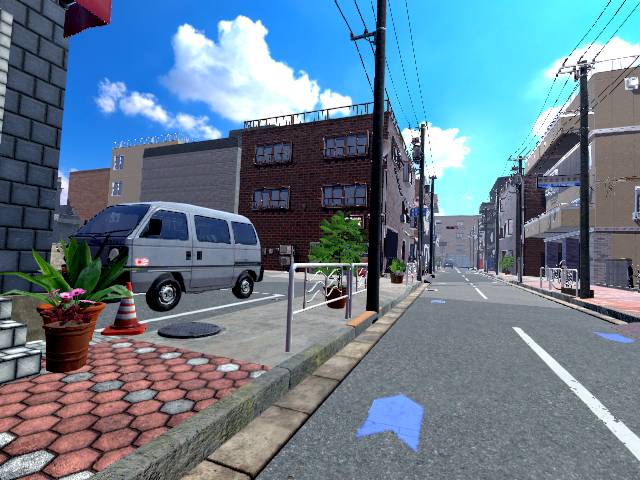} &
\includegraphics[width=\linewidth]{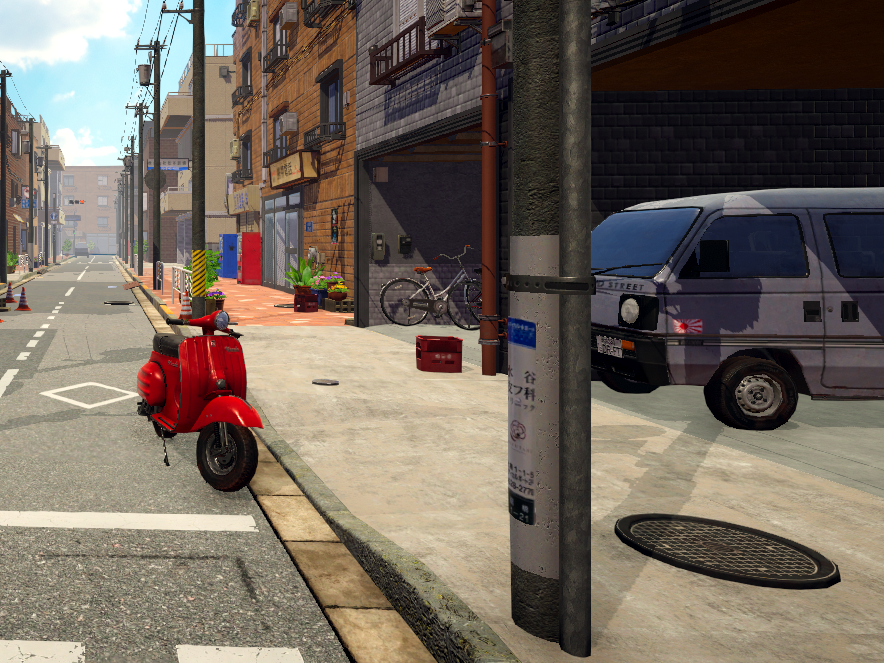} &
\includegraphics[width=\linewidth]{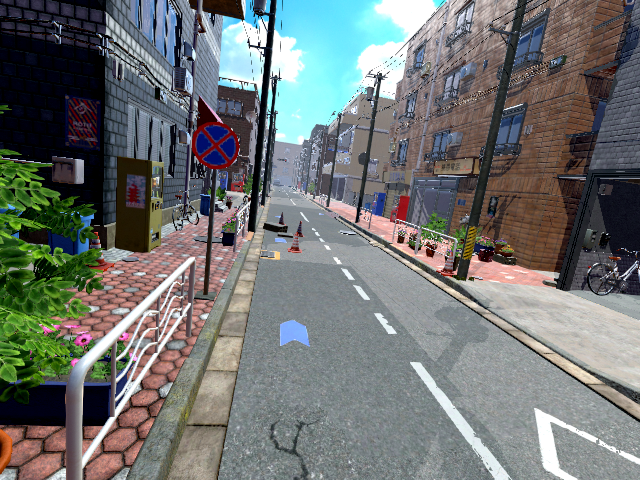} \tabularnewline
\tabularnewline
\end{tabular}
 \vspace{-1ex}
\caption{\label{fig:sim_environments} \textbf{Simulation Environments}. The selected environments are challenging since they contain obstacles with complex geometry or narrow passages(\textbf{C}).
Despite training only in simulated forests~(\textbf{A}) and artificial environments with obstacles of convex shapes~(\textbf{B}), our approach generalizes zero-shot to complex disaster zones~(\textbf{D}) and urban environments~(\textbf{E}).
An illustration of the simulation environments can also be found in \href{https://www.youtube.com/watch?v=ALs60ij8JA8}{Movie S2}.
}
\end{figure*}

\begin{figure*}
    \centering
    \includegraphics[width=0.8\textwidth]{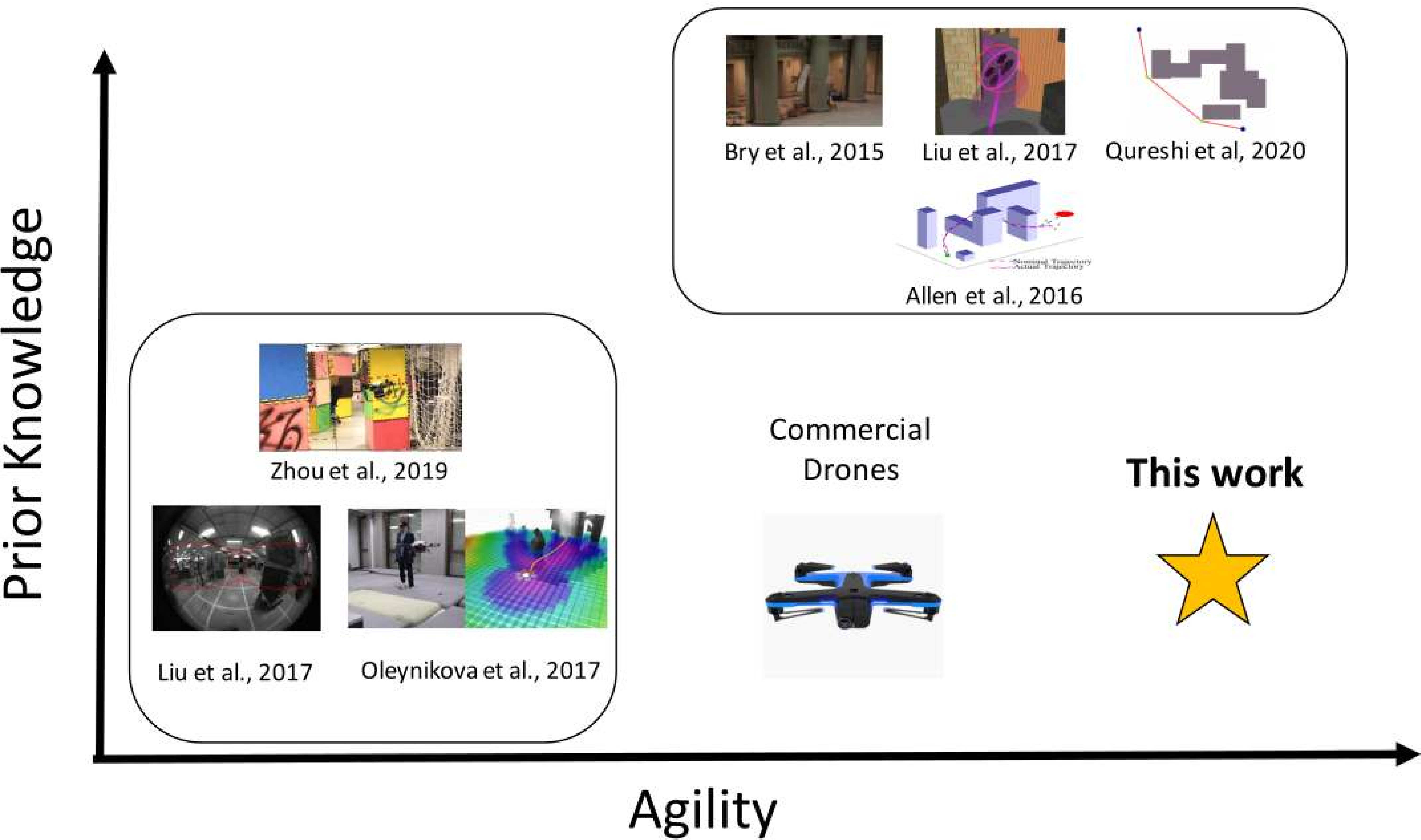}
    \caption{\textbf{Taxonomy of existing approaches for drone navigation} in challenging and cluttered environments. Approaches are ordered with respect to the required prior knowledge about the environment and the maximum agility they achieve.}
    \label{fig:pc:map_rel_work}
\end{figure*}

\begin{figure*}
\centering
\begin{subfigure}{.5\textwidth}
  \centering
  \begin{tabular}{c}
    \textbf{(a)} Without global planning \\
    \includegraphics[trim=400 200 100 200,clip,width=.9\linewidth]{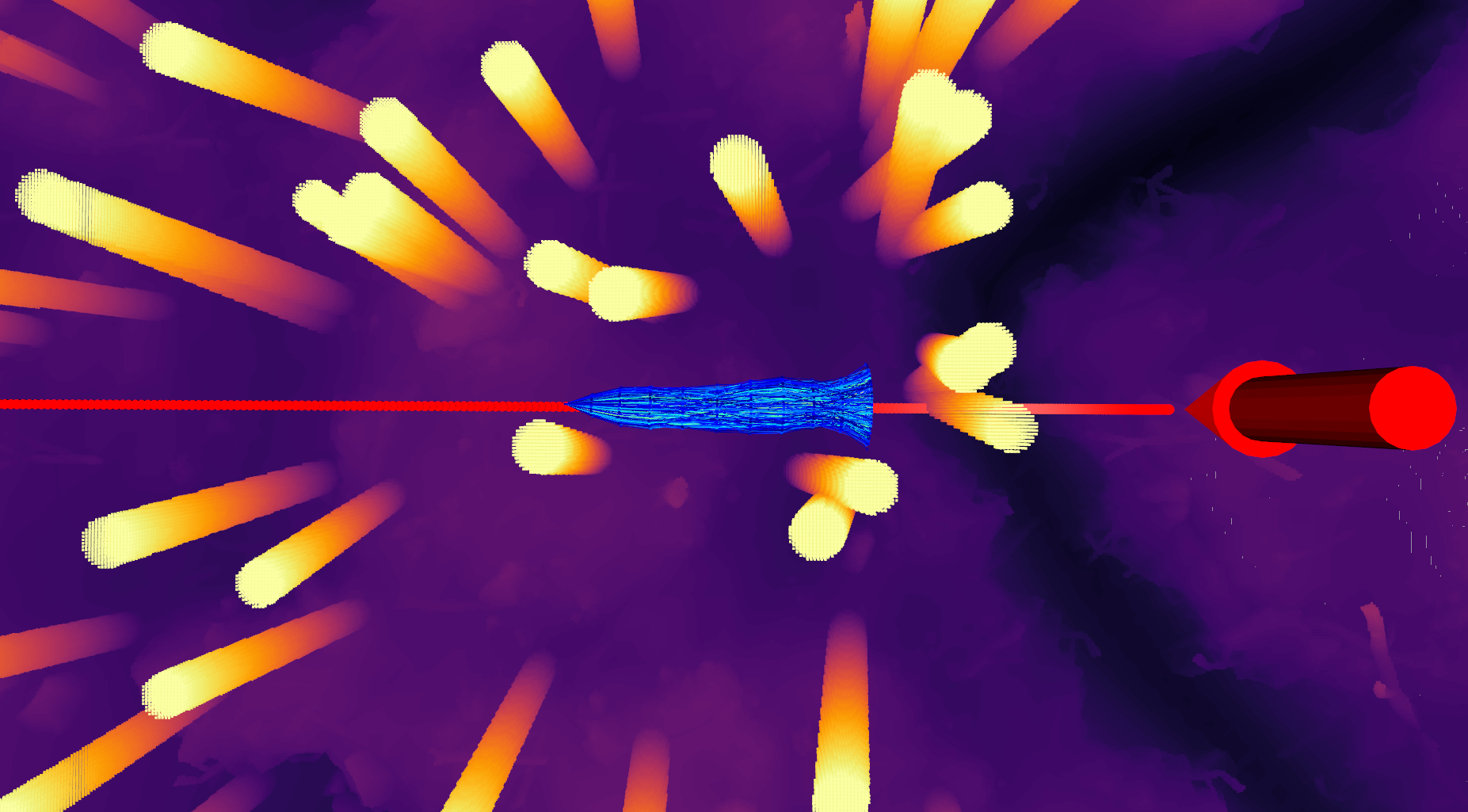} \\
  \end{tabular}
\end{subfigure}%
\begin{subfigure}{.5\textwidth}
  \centering
  \begin{tabular}{c}
    \textbf{(b)} With global planning \\
    \includegraphics[trim=400 200 100 200,clip,width=.9\linewidth]{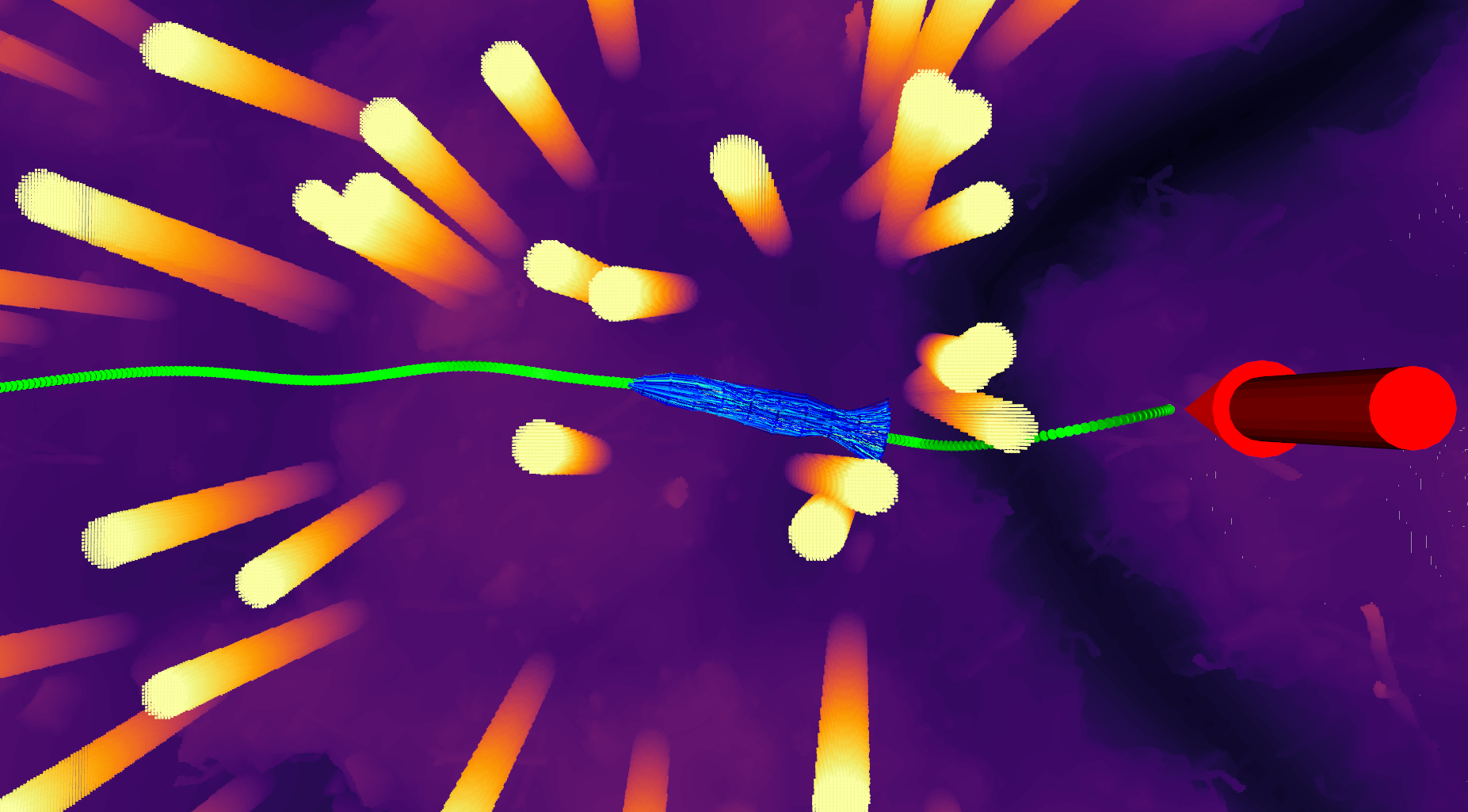} \\
  \end{tabular}
\end{subfigure}%
\caption{\textbf{Influence of global planning on the sampled trajectories}. Sampling around the raw reference trajectory (in red) strictly limits the expert's sight to the immediate horizon (a). Conversely, sampling around a global collision-free trajectory (in green) results in a bias toward obstacle-free regions even beyond the immediate horizon (b). Best viewed in color.}
\label{fig:global_plan_helps}
\end{figure*}

\end{document}